\documentclass[10pt,twocolumn,letterpaper]{article}

\usepackage[pagenumbers]{cvpr} % To force page numbers, e.g. for an arXiv version

\usepackage{graphicx}
\usepackage{amsmath}
\usepackage{amssymb}
\usepackage{booktabs}
\usepackage{bm}
\usepackage[accsupp]{axessibility}  % Improves PDF readability for those with disabilities.
\usepackage{color}

\definecolor{turquoise}{cmyk}{0.65,0,0.1,0.3}
\definecolor{purple}{rgb}{0.65,0,0.65}
\definecolor{dark_green}{rgb}{0, 0.5, 0}
\definecolor{orange}{rgb}{0.8, 0.6, 0.2}
\definecolor{red}{rgb}{0.8, 0.2, 0.2}
\definecolor{darkred}{rgb}{0.6, 0.1, 0.05}
\definecolor{blueish}{rgb}{0.0, 0.3, .6}
\definecolor{light_gray}{rgb}{0.7, 0.7, .7}
\definecolor{pink}{rgb}{1, 0, 1}
\definecolor{greyblue}{rgb}{0.25, 0.25, 1}

\newcommand{\SO}[1]{\bm{\mathrm{SO}(#1)}}

\usepackage[pagebackref,breaklinks,colorlinks]{hyperref}
\usepackage{caption}

\newcommand{\insertfig}{
    \includegraphics[width=\linewidth]{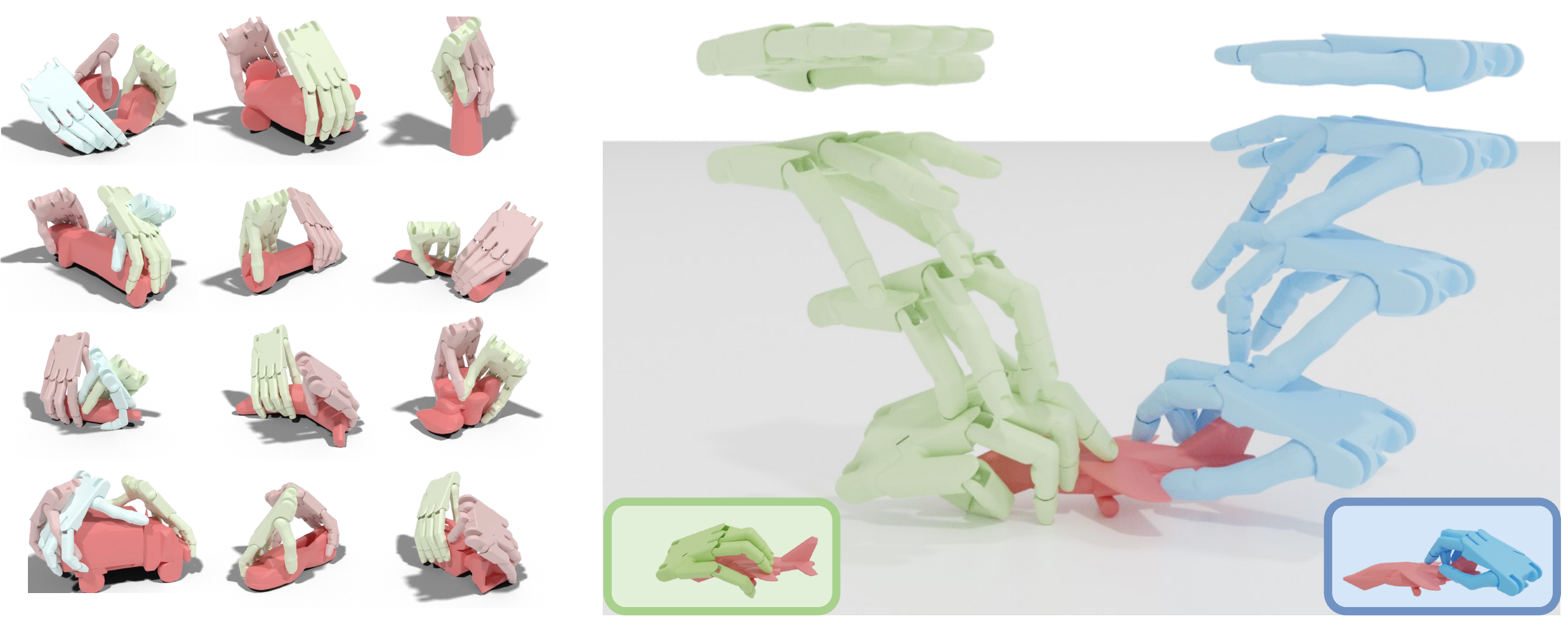}
    \vspace{-2em}
    \captionof{figure}{\textbf{UniDexGrasp via grasp proposal generation and goal-conditioned execution.} Left (\textit{grasp proposals}): each figure demonstrates two or three diverse and high-quality grasp proposals that vary greatly in rotation, translation, and joint angles; right (\textit{grasp execution}): given a grasp goal pose, our highly generalizable goal-conditioned grasping policy can grasp the object in the way specified by the goal, as shown in the green and blue trajectories and their corresponding goals.}
    \vspace{1em}
}
\makeatletter
\apptocmd{\@maketitle}{\centering\insertfig}{}{}  % insert the figure after authors
\makeatother

\usepackage[capitalize]{cleveref}
\crefname{section}{Sec.}{Secs.}
\Crefname{section}{Section}{Sections}
\Crefname{table}{Table}{Tables}
\crefname{table}{Tab.}{Tabs.}

\def\cvprPaperID{7986} % *** Enter the CVPR Paper ID here
\def\confName{CVPR}
\def\confYear{2022}

\begin{document}

%%%%%%%%% TITLE - PLEASE UPDATE
%\title{Propose How Then Grasp: Learning Universal Policy\\
% for Diverse Dexterous Robotic Grasping from Point Clouds}

\title{UniDexGrasp: Universal Robotic Dexterous Grasping\\
via Learning Diverse Proposal Generation and Goal-Conditioned Policy}

\author{
% For a paper whose authors are all at the same institution,
% omit the following lines up until the closing ``}''.
% Additional authors and addresses can be added with ``\and'',
% just like the second author.
% To save space, use either the email address or home page, not both
Yinzhen Xu\footnotemark[1] \textsuperscript{ 1,2,3},
\quad Weikang Wan\footnotemark[1] \textsuperscript{ 1,2},
\quad Jialiang Zhang\footnotemark[1] \textsuperscript{ 1,2},
\quad Haoran Liu\footnotemark[1] \textsuperscript{ 1,2},
\quad Zikang Shan\textsuperscript{1}, \\
\quad Hao Shen\textsuperscript{1},
\quad Ruicheng Wang\textsuperscript{1},
\quad Haoran Geng\textsuperscript{1,2},
\quad Yijia Weng\textsuperscript{4},
\quad Jiayi Chen\textsuperscript{1},\\
\quad Tengyu Liu\textsuperscript{3},
\quad Li Yi\textsuperscript{5},
\quad He Wang\footnotemark[2] \textsuperscript{ 1,2}\\
\textsuperscript{1} Center on Frontiers of Computing Studies, Peking University
\quad\textsuperscript{2} School of EECS, Peking University \\
\textsuperscript{3} Beijing Institute for General AI 
\quad\textsuperscript{4} Stanford University
\quad\textsuperscript{5} Tsinghua University \\
\href{https://pku-epic.github.io/UniDexGrasp/}{https://pku-epic.github.io/UniDexGrasp/}
}
\maketitle
\renewcommand{\thefootnote}{\fnsymbol{footnote}}
\footnotetext[1]{Equal contribution.}
\footnotetext[2]{Corresponding author.}

%%%%%%%%% TEASER

%%%%%%%%% ABSTRACT
\begin{abstract}
In this work, we tackle the problem of learning universal robotic dexterous grasping from a point cloud observation under a table-top setting.
The goal is to grasp and lift up objects in high-quality and diverse ways and generalize across hundreds of categories and even the unseen.
Inspired by successful pipelines used in parallel gripper grasping, we split the task into two stages: 1) grasp proposal (pose) generation and 2) goal-conditioned grasp execution. For the first stage, we propose a novel probabilistic model of grasp pose conditioned on the point cloud observation that factorizes rotation from translation and articulation. 
Trained on our synthesized large-scale dexterous grasp dataset, this model enables us to sample diverse and high-quality dexterous grasp poses for the object point cloud.
For the second stage, we propose to replace the motion planning used in parallel gripper grasping with a goal-conditioned grasp policy, due to the complexity involved in dexterous grasping execution. Note that it is very challenging to learn this highly generalizable grasp policy that only takes realistic inputs without oracle states. We thus propose several important innovations, including state canonicalization, object curriculum, and teacher-student distillation.
Integrating the two stages, our final pipeline becomes the first to achieve universal generalization for dexterous grasping, demonstrating an average success rate of more than 60\% on thousands of object instances, which significantly outperforms all baselines, meanwhile showing only a minimal generalization gap.  
% When integrating the two stages together, our final pipeline, for the first time, shows universal dexterous grasping on thousands of object instances with more than 60\% success rate  and significantly outperforms all baselines. Our experiments show a minimal generalization gap between the seen and unseen instances, further demonstrating the universality of our method.
   %execute the static grasp prediction using a goal-conditioned reinforcement learning algorithm\todo{?}. Such an intuitive divide-and-conquer paradigm is demonstrated to boost learning efficiency and improve the success rate. Moreover, because the input only relies on a point cloud fused from depth images and the robot's proprioception, 
\end{abstract}

%%%%%%%%% BODY TEXT
\vspace{-3mm}
\section{Introduction}
\label{sec:intro}

Robotic grasping is a fundamental capability for an agent to interact with the environment and serves as a prerequisite to manipulation, which has been extensively studied for decades.
Recent years have witnessed great progress in developing grasping algorithms for parallel grippers \cite{sundermeyer2021contact,breyer2021volumetric,fang2020graspnet,gou2021RGB,Wang_2021_ICCV,fang2022transcg} that carry high success rate on universally grasping unknown objects. However, one fundamental limitation of parallel grasping is its low dexterity which limits its usage to complex and functional object manipulation.

Dexterous grasping provides a more diverse way to grasp objects and thus is of vital importance to robotics for functional and fine-grained object manipulation~\cite{akkaya2019solvingrubik, qin2021dexmv, qingeneralizable, wu2022learning,liu2022hoi4d}.
However, the high dimensionality of the actuation space of a dexterous hand is both the advantage that endows it with such versatility and the major cause of the difficulty in executing a successful grasp.
As a widely used five-finger robotic dexterous hand, ShadowHand~\cite{shadowhand} amounts to 26 degrees of freedom (DoF), in contrast with 7 DoF for a typical parallel gripper. Such high dimensionality magnifies the difficulty in both generating valid grasp poses and planning the execution trajectories, and thus distinguishes the dexterous grasping task from its counterpart for parallel grippers.
Several works have tackled the grasping pose synthesis problem\cite{GraspIt, tengyuDifferentiableForceClosure, sundermeyer2021contact, contactDB}, however, they all assume oracle inputs (full object geometry and states). Very few works\cite{qingeneralizable, chenlearning} tackle dexterous grasping in a realistic robotic setting, but so far no work yet can demonstrate universal and diverse dexterous grasping that can well generalize to unseen objects. 

%Another unneglectable factor is the lack of large-scale dexterous grasping labels, including datasets for static grasps that feature both diversity and physical plausibility, and dynamic grasping trajectories that reinforcement learning algorithms usually require. In previous endeavors, one way to obtain the grasp poses for learning is either predictions from human-object interactions, like in~\cite{mandikal2021dexvip}\todo{what else?}, which suffer from low accuracy, or recorded data by costly motion capture equipment \todo{like in ?}. Alternatively, synthetic data can be used, \todo{introduce the dataset?}.
In this work, we tackle this very challenging task: learning universal dexterous grasping skills that can generalize well across hundreds of seen and unseen object categories in a realistic robotic setting and only allow us to access depth observations and robot proprioception information. Our dataset contains more than one million grasps for 5519 object instances from 133 object categories, which is the largest robotic dexterous grasping benchmark to evaluate universal dexterous grasping.

Inspired by the successful pipelines from parallel grippers, we propose to decompose this challenging task into two stages:
1) \textbf{dexterous grasp proposal generation}, in which we predict diverse grasp poses given the point cloud observations;
and 2) \textbf{goal-conditioned grasp execution}, in which we take one grasp goal pose predicted by stage 1 as a condition and generates physically correct motion trajectories that comply with the goal pose. Note that both of these two stages are indeed very challenging, for each of which we contribute several innovations, as explained below.

For dexterous grasp proposal generation, we devise a novel conditional grasp pose generative model that takes point cloud observations and is trained on our synthesized large-scale table-top dataset. 
Here our approach emphasizes the diversity in grasp pose generation, since the way we humans manipulate objects can vary in many different ways and thus correspond to different grasping poses. Without diversity, it is impossible for the grasping pose generation to comply with the demand of later dexterous manipulation. 
Previous works~\cite{jiang2021graspTTA} leverages CVAE to jointly model hand rotation, translation, and articulations and we observe that such CVAE suffers from severe mode collapse and can't generate diverse grasp poses, owing to its limited expressivity when compared to conditional normalizing flows~\cite{dinh2014nice,dinh2016density,kingma2018glow,falorsi2019reparameterizing,papamakarios2021normalizing} and conditional diffusion models~\cite{rombach2022high,saharia2022photorealistic,batzolis2021conditional}. 
However, no works have developed normalizing flows and diffusion models that work for the grasp pose space, which is a Cartesian product of $\SO3$ of hand rotation and a Euclidean space of the translation and joint angles. We thus propose to decompose this conditional generative model into two conditional generative models: a conditional rotation generative model, \textit{namely} GraspIPDF, leveraging ImplicitPDF~\cite{implicitpdf2021} (in short, IPDF) and a conditional normalizing flow,  \textit{namely} GraspGlow, leveraging Glow~\cite{kingma2018glow}. Combining these two modules, we can sample diverse grasping poses and even select what we need according to language descriptions. The sampled grasps can be further refined to be more physically plausible via ContactNet, as done in ~\cite{jiang2021graspTTA}.

For our grasp execution stage, we learn a goal-conditioned policy that can grasp any object in the way specified by the grasp goal pose and only takes realistic inputs: the point cloud observation and robot proprioception information, as required by real robot experiments.
Note that reinforcement learning (RL) algorithms usually have difficulties with learning such a highly generalizable policy, especially when the inputs are visual signals without ground truth states.
To tackle this challenge, we leverage a teacher-student learning framework that first learns an oracle teacher model that can access the oracle state inputs and then distill it to a student model that only takes realistic inputs. Even though the teacher policy gains access to oracle information, making it successful in grasping thousands of different objects paired with diverse grasp goals is still formidable for RL. We thus introduce two critical innovations: a canonicalization step that ensures $\SO2$ equivariance to ease the policy learning; and an object curriculum that first learns to grasp one object with different goals, then one category, then many categories, and finally all categories. 

Extensive experiments demonstrate the remarkable performance of our pipelines. In the grasp proposal generation stage, our pipeline is the only method that exhibits high diversity while maintaining the highest grasping quality. The whole dexterous grasping pipeline, from the vision to policy, again achieves impressive performance in our simulation environment and, for the first time, demonstrates a universal grasping policy with more than 60\% success rate and remarkably outperforms all the baselines. We will make the dataset and the code publicly available to facilitate future research.

%directly consuming visual inputs and 
%We propose a generative model trained on a large-scale dataset containing diverse grasp poses to predict multiple grasp proposals for every single target object. Then a grasp pose proposal without collision with the scene is selected as our goal pose. To execute the grasp, inspired by~\cite{christen2022dgrasp}\todo{?}, we train a policy network conditioned on the goal to hold and lift the object.
%For better generalizability to novel objects and transferability to real world, our method only requires a point cloud fused from depth images and the robot's proprioception. We conduct experiments in Isaac Gym~\cite{isaacgym} to provide evidence for the generalizability of our method.
    
% \begin{figure*}[h]
%     \centering
%     \includegraphics[width=300pt]{fig/teaser_1.png}
%     \caption{Caption}
%     \label{fig:teaser-middle}
% \end{figure*}
\section{Related Work}
\noindent\textbf{Dexterous Grasp Synthesis}\quad 
Dexterous grasp synthesis, a task aiming to generate valid grasping poses given the object mesh or point cloud, falls into two categories. 
First, non-learning methods serve to generate large synthetic datasets. Among which, \textit{GraspIt!}~\cite{GraspIt} is a classical tool that synthesizes stable grasps by collision detection, commonly adopted by early works~\cite{liu2020deep,kaixu2020highDofGrasp,brahmbhatt2019contactgrasp}. Recently, an optimization-based method~\cite{tengyuDifferentiableForceClosure} greatly improves grasp diversity with its proposed differentiable force closure estimator, enabling datasets of higher quality~\cite{wang2022dexgraspnet,li2022gendexgrasp}. 
Second, learning-based methods~\cite{jiang2021graspTTA,grady2021contactopt,brahmbhatt2019contactgrasp,synthesizingGraspConfigurationsWithSpecifiedContactRegions,corona2020ganhand,liu2020deep} learn from these datasets to predict grasps with feed-forward pipelines, but struggle to possess quality and diversity at the same time. To tackle this problem, we propose to build a conditional generative model that decouples rotation from translation and articulation.

\noindent\textbf{Probabilistic Modeling on $\SO3\times \mathbb{R}^n$} \quad%\todo{rewrite}}\quad 
One way to generate diverse grasping poses is to learn the distribution of plausible poses using ground truth poses. 
%Hand poses lie in $\SO3\times \mathbb{R}^n$. \todo{elaborate how people modeled so3 and rn, and describe how we did it}
GraspTTA~\cite{jiang2021graspTTA} uses conditional VAE and suffers from severe model collapse which leads to limited diversity. In contrast, normalizing flow is capable of modeling highly complex distributions as it uses negative log-likelihood (NLL) as loss, suiting our needs for grasping proposal generation. 
% Intuitively, normalizing flow is a bijection $f: \mathbb{R}^D\rightarrow\mathbb{R}^D$ which can relocate samples from a simple base distribution $p_u(u)$, typically the standard normal distribution $N(0, I)$, to form the target distribution $p_x(x)$. By using normalizing flow, one can directly calculate the probability $p_x(x)=p_u(u)|\det{f}|^{-1}$ and also sample from the predicted distribution efficiently, as shown in Sec.\ref{sec:flow-sampling}. 
% There have been lots of works focusing on building normalizing flow in the Euclidean space, such as NICE\cite{dinh2014nice}, RealNVP\cite{dinh2016density}, and Glow\cite{kingma2018glow}. In their work, coupling layer is used to improve the expressivity. 
Building normalizing flow (NF) in the Euclidean space has been well studied~\cite{dinh2014nice,dinh2016density,kingma2018glow}. 
But unfortunately, hand pose consists of a $\SO3$ part (the rotation) and a $\mathbb{R}^n$ part (the translation and joint angles), and using normalizing flow in $\SO3$ is hard due to its special topological structure.  
Relie~\cite{falorsi2019reparameterizing} and ProHMR~\cite{kolotouros2021probabilistic} both perform NF in the Euclidean space and then map it into $\SO3$. As these two spaces are not topologically equivalent, they suffer from discontinuity and infinite-to-one mapping respectively. A more detailed analysis of these phenomena is in {\color{purple}Sec. B.1.2} of our supp. 
%ReLie~\cite{falorsi2019reparameterizing} performs normalizing flow in $\mathbb{R}^3$ and view the Euclidean vector as the axis-angle representation of $\SO3$. However, this practice introduces discontinuity to the flow as these two spaces are not topologically equivalent.
%ProHMR~\cite{kolotouros2021probabilistic} performs normalizing flow in $\mathbb{R}^6$ and uses the Gram-Schmidt orthogonalization to get 6D representation of $\SO3$. Although it doesn't suffer from discontinuity, this orthogonalization is an infinite-to-one mapping and is thus harmful to normalizing flow training.
In comparison, IPDF~\cite{implicitpdf2021} uses a neural network to output unnormalized log probability and then normalize it with uniform samples or grids on $\SO3$ and also uses NLL as loss. As IPDF is insensitive to topological structure, it is a better choice to model distributions on $\SO3$ than existing NFs.
% ReLie\cite{falorsi2019reparameterizing}, a method to perform normalizing flow on $\SO3$, can be briefly described as performing normalizing flow on $\mathbb{R}^3$ and applying the exponential map to transform a Euclidean vector into an element of $\SO3$. 
% However, as this method equals to using the axis angle representation, it suffers from discontinuity.
Therefore, our grasp proposal module decouples rotation from translation and joint angles and models these distributions separately with IPDF~\cite{implicitpdf2021} and Glow~\cite{kingma2018glow}. 
\label{sec:flow-related-work}

\noindent\textbf{Dexterous Grasp Execution}
% Due to anthropomorphism, dexterous hands have received extensive attention for its potential of human-like manipulation. 
% Dexterous manipulation is of high potential yet very challenging due to its high dexterity. 
Executing a dexterous grasp requires an agent to perform a complete trajectory, rather than a static grasping pose. Previous approaches have used analytical methods~\cite{baiyunfeiKarenLiu, pushgrasping, andrews2013goal} to model hand and object kinematics and dynamics, and then optimized trajectories for robot control. However, these methods typically require simplifications such as using simple finger and object geometries to make planning tractable.
More recently, reinforcement and imitation learning techniques have shown promise for dexterous grasping~\cite{qin2021dexmv,christen2022dgrasp,rajeswaran2017learning,mandikal2021graff,she2022learning, wu2022learning}. However, these methods rely on omniscient knowledge of the object mesh and struggle to handle realistic task settings, making them unsuitable for deployment in the real world.
To address this issue, recent works have explored using raw RGB images~\cite{mandikal2021dexvip,mandikal2021graff} or 3D point clouds~\cite{qin2022dexpoint}  as policy inputs. However, none of these methods have been able to generalize to a large number of objects under raw vision input. In contrast, our goal-conditioned grasp execution method achieves universal generalization on thousands of object instances by leveraging a teacher-student distillation trick, object curriculum learning, and state canonicalization.

\begin{figure*}[h]
    \centering
    \includegraphics[width=1\linewidth]{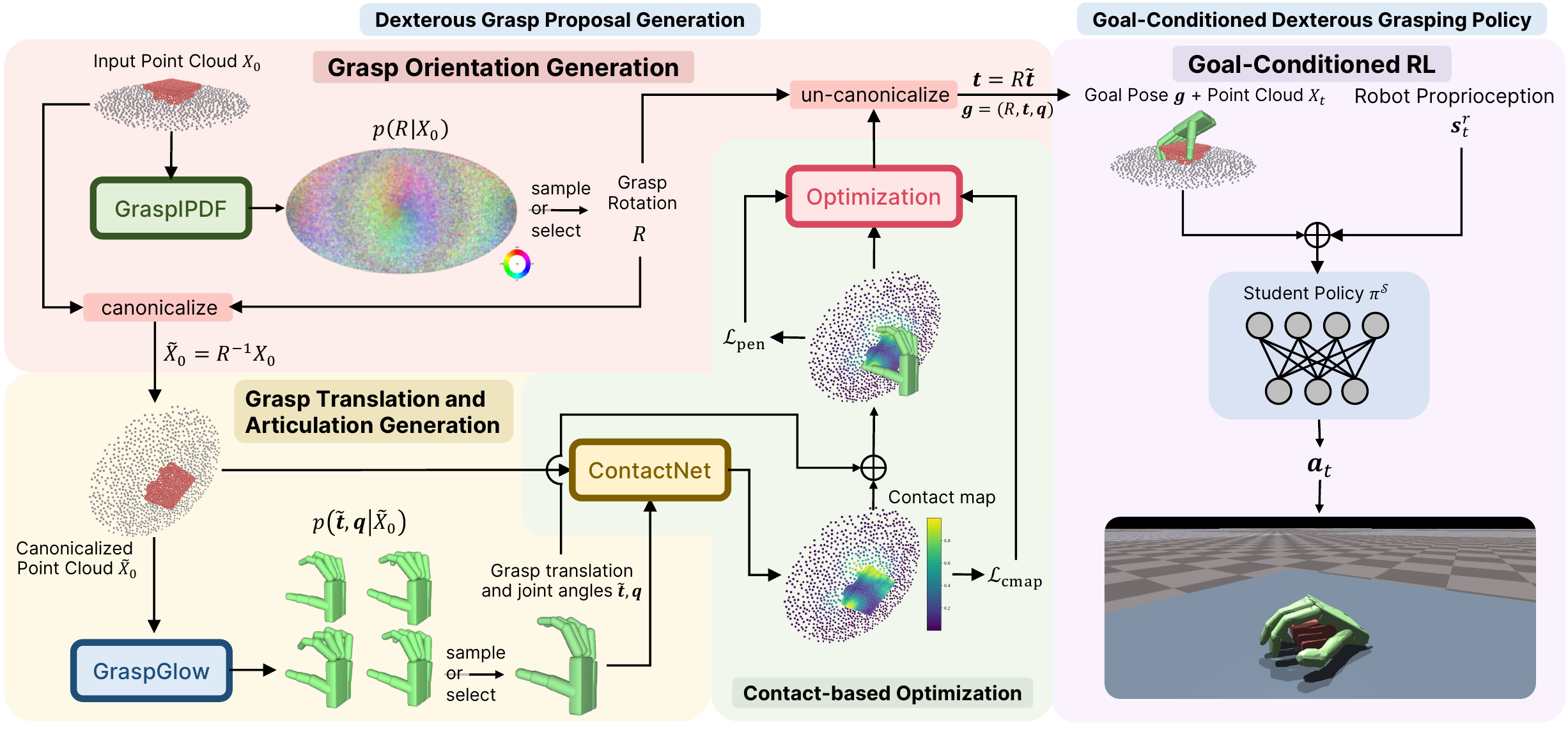}
    \caption{\textbf{Method overview}. The left part is the first stage, which generates a dexterous grasp proposal. The input is the object point cloud at time step 0, $X_0$, fused from depth images, with ground truth segmentation of the table and the object. A rotation $R$ is sampled from the distribution implied by the GraspIPDF, and the point cloud will be canonicalized by $R^{-1}$ to $\tilde{X}_0$. The GraspGlow then samples the translation $\tilde{\bm{t}}$ and joint angles $\bm{q}$. Next, the ContactNet takes $\tilde{X}_0$ and a point cloud $\tilde{X}_H$ sampled from the hand to predict the ideal contact map $\bm{c}$ on the object. Then, the predicted hand pose is optimized based on the contact information. The final goal pose is transformed by $R$ to align with the original visual observation. The right part is the second stage, the goal-conditioned dexterous grasping policy that takes the goal $\bm{g}$, point cloud $X_t$ and robot proprioception $\bm{s}^r_t$ to take actions accordingly.
    %\todo{todos in comment}%\todo{input point cloud $X_0$, RL point cloud $X^S_t$, robot proprioception should also have a t subscript, the r subscript is not necessary}~\todo{on top of the figure, add titles "proposal generation" "execution"}~\todo{explain how the goal pose $g=(R,t,q)$ for RL is sampled}~\todo{$t$ should have a tilde, and explain $t=un-cannonicalized(\tilde t)$}~\todo{$p(t,q|X,R)\to p(\tilde t,q|\tilde X)$}
    }
    \label{fig:MainPipeline}
\vspace{-3mm}
\end{figure*}
\section{Method}
We propose UniDexGrasp, a two-stage pipeline for generalizable dexterous grasping. We divide the task into two phases: 1) grasp proposal generation (Sec.~\ref{sec:dexterous_grasp_proposal_generation}), 2) goal-conditioned grasp execution (Sec.~\ref{sec:dexterous_grasping_policy}). First, the grasp proposal generation module takes the object point cloud and samples a grasp proposal. Then, the goal-conditioned grasping policy takes this proposal as goal pose, and executes the grasp in the Isaac Gym simulator~\cite{isaacgym}, taking only point cloud observations and robot proprioception as input in every time step $t$. 

%\todo{leave enough vspace for sections}

% \vspace{-4mm}
\subsection{Problem Settings and Method Overview}
% \todo{first paragraph, define grasp proposal, explain the inputs and outputs clearly, do not distribute the settings into the other subsections, the rest}

The grasp proposal generation module, shown on the left part of Fig.~\ref{fig:MainPipeline}, takes the object and table cloud $X_0\in \mathbb{R}^{N\times 3}$ as input, and samples a grasp proposal $\boldsymbol g=(R,\boldsymbol t,\boldsymbol q)$ out of a distribution, where $R\in \SO3,\boldsymbol t\in \mathbb{R}^3,\boldsymbol q\in \mathbb{R}^K$ represent the root rotation, root translation, and joint angles of the dexterous hand, and $K$ is the total degree-of-freedom of the hand articulations. The proposed grasp $\boldsymbol g$ will be the goal pose of the next module, shown on the right part of Fig.~\ref{fig:MainPipeline}. 

% \todo{second paragraph, define grasp execution, explain the inputs and outputs clearly}

The final goal-conditioned grasp execution module is a vision-based policy that runs in the IsaacGym~\cite{isaacgym} physics simulator. In each time step $t$, the policy takes the goal pose $\boldsymbol g$ from the previous module, object and the scene point cloud $X_t$, and robot proprioception $\boldsymbol s_t^r$ as observation, and outputs an action $\boldsymbol a_t$. The policy should work across different object categories and even unseen categories. To simplify the problem, we initialize the hand with an initial translation $\bm{t}_0 = (0, 0, h_0)$ and an initial rotation $R_0 = (\frac{\pi}{2}, 0, \phi_0)$, where $h_0$ is a fixed height and the hand rotation is initialized so that the hand palm faces down and its $\phi_0 =\phi$. The joint angles of the hand are set to zero.
The task is to grasp the object as specified by the goal grasp label $g$ and lift it to a certain height. The task is successful if the position difference between the object and the target point is smaller than the threshold value $\bm{t}_0$.

Since directly training such a vision-based policy using reinforcement learning is challenging, we use the idea of \textit{teacher-student learning}.
We first use a popular on-policy RL algorithm, PPO\cite{schulman2017proximal}, with our proposed object curriculum learning and state canonicalization, to learn an oracle teacher policy that can access the ground-truth states of the environment (\eg object poses, velocities and object full point cloud).
This information is very useful to the task and available in the simulator but not in the real world. 
Once the teacher finishes training, we use an imitation learning algorithm, DAgger\cite{ross2011reduction}, to distill this policy to a student policy that can only access realistic inputs.

% \todo{explain side by side with fig 2, use mathematical symbols to refer to the input and outputs of submodules, make all the symbols consistent across all sections, rigorously clarify the notations and make the section shorter, move the motivations to intro}

\subsection{Dexterous Grasp Proposal Generation}\label{sec:dexterous_grasp_proposal_generation}
% \vspace{-2mm}
% \subsubsection{Overview and Notations} 

In this subsection, we introduce how we model the conditional probability distribution $p(\bm{g}|X_0): \SO3\times \mathbb{R}^{3+K} \rightarrow \mathbb{R}$. By factorizing $p(\bm{g}|X_0)$ into two parts $p(R|X_0)\cdot p(\bm{t}, \bm{q}|X_0, R)$, we propose a three-stage pipeline: 1) given the point cloud observation, predict the conditional distribution of hand root rotation $p(R|X_0)$ using GraspIPDF (see Sec.~\ref{sec:graspipdf}) and then sample a single root rotation; 2) given the point cloud observation and the root rotation, predict the conditional distribution of the hand root translation and joint angles $p(\bm{t}, \bm{q}|X_0, R) = p(\tilde{\bm{t}}, \bm{q}|\tilde{X}_0)$ using GraspGlow (see Sec.~\ref{sec:graspglow}) and then sample a single proposal; 3) optimize the sampled grasp pose $g$ with ContactNet to improve physical plausibility (see Sec.~\ref{sec:tta}).

\vspace{-3mm}
\subsubsection{GraspIPDF: Grasp Orientation Generation}
\vspace{-1mm}
\label{sec:graspipdf}
% \quad The original IPDF~\cite{implicitpdf2021} is a probabilistic model over $\SO3$ that maps a pair of input and rotation, \ie, $(\mathcal{X}, R)$, where the input $\mathcal{X}$ in IPDF is an image, to an unnormalized joint log probability that indicates the likelihood of this pair.
\quad Inspired by IPDF~\cite{implicitpdf2021}, a probabilistic model over $\SO3$, 
we propose GraspIPDF $f(X_0, R)$ to predict the conditional probability distribution $p(R|X_0)$ of the hand root rotation $R$ given the point cloud observation $X_0$. This model takes $X_0$ and $R$ as inputs, extracts the geometric features with a PointNet++~\cite{qi2017pointnetplusplus} backbone, and outputs an unnormalized joint log probability density $f(X_0, R)  = \alpha\, \mathrm{log}(p(X_0,R))$, where $\alpha$ is a normalization constant. The normalized probability density is recovered by computing:
\begin{equation}
p(R|X_0) = \cfrac{p(X_0, R)}{p(X_0)} \approx \cfrac{1}{V} \cfrac{\mathrm{exp}(f(X_0, R))}{\sum^M_i \mathrm{exp}(f(X_0, R_i))}
\end{equation}
where $M$ is the number of volume partitions and $V=\cfrac{\pi^2}{M}$ is the volume of partition.
% \quad This grasp orientation generation module predicts a conditional probability distribution $p(R|X_0)$ of the hand root rotation $R$ over $\SO3$, given the point cloud observation $X$.

% Inspired by IPDF~\cite{implicitpdf2021}, a probabilistic model over $\SO3$ that maps a pair of input and rotation, \ie, $(\mathcal{X}, R)$, where the input $\mathcal{X}$ in IPDF is an image, to an unnormalized joint log probability that indicates the likelihood of this pair, 
% we propose GraspIPDF $f(X, R)$, which takes the point cloud observation $X_0$ and a rotation $R$ as inputs, and outputs an unnormalized joint log probability density $f(X, R)  = \alpha\, \mathrm{log}(p(X,R))$,
% where $\alpha$ is a constant. 
% In GraspIPDF, we use a PointNet~\cite{qi2016pointnet} for geometric feature extraction and keep the rest of the architecture similar to IPDF. To obtain the conditional probability density, we need to compute the normalization constant
% we have $p(R|X) = \cfrac{p(X, R)}{p(X)} \approx \cfrac{1}{V} \cfrac{\mathrm{exp}(f(X, R))}{\sum^N_i \mathrm{exp}(f(X, R_i))}$, where $N$ is the number of volume partitions and $V=\cfrac{\pi^2}{N}$ is the volume of partition. 

During train time, GraspIPDF is supervised by an NLL loss $\mathcal{L} = -\log(p(R_0 | X_0))$, where $R_0$ is a ground-truth hand root rotation. During test time, we generate an equivolumetric grid on $\SO3$ as in~\cite{implicitpdf2021,hopffibration,Gorski_2005} and sample rotations according to their queried probabilities.

\vspace{-3mm}
\subsubsection{GraspGlow: Grasp Translation and Articulation Generation given Orientation}
\label{sec:graspglow}
\quad To condition a probabilistic model on $X_0$ and $R$ simultaneously, we propose to canonicalize the point cloud to $\tilde{X}_0= R^{-1}X_0$. This trick simplifies the task from predicting valid grasps for observation $X_0$ with $R$ as the hand root rotation, to predicting it for $\tilde{X}_0$ with identity as hand rotation. Then, we use a PointNet~\cite{qi2016pointnet} to extract the features of $\tilde{X}_0$, and model the conditional probability distribution $p(\bm{t},\bm{q}|X_0,R)=p(\tilde{\bm{t}},\bm{q}|\tilde{X}_0)$ where $\tilde{\bm{t}}=R^{-1}\bm{t}$ with Glow~\cite{kingma2018glow}, a popular normalizing flow model that handles probabilistic modeling over Euclidean spaces. 

During train time, the model is supervised by an NLL loss $\mathcal{L}_{\text{NLL}}=-\log(p(\tilde{\bm{t}}_{\textrm{gt}},\bm{q}_{\textrm{gt}}|\tilde{X}_0)) $ using ground truth grasp data tuples $(X_0,R_{\textrm{gt}},\bm{t_{\textrm{gt}}},\bm{q_{\textrm{gt}}})$. During test time, samples are drawn from the base distribution of Glow, and reconstructed into grasp poses using the bijection of the normalizing flow. 
% \noindent\textbf{Loss} Similar to GraspIPDF, we minimize the negative log-likelihood (NLL) of the ground-truth hand root translation and joint angles using the canonicalized point cloud as the condition: $\mathcal{L}_{\text{NLL}}=-\log(p(\tilde{\bm{t}}_\text{gt},\bm{q}_\text{gt}|R_\text{gt}^{-1}X)) $

% Through this training, GraspGlow can fit the data distribution and thus provide diverse samples, however, the samples might be physically implausible due to the lack of constraint. Also, the performance may drop at test time as it is trained using ground-truth rotation and tested using the GraspIPDF's samples. To alleviate those problems, we train our model end-to-end in a second stage, see Sec.\ref{sub:subsection_end_to_end}.

% \noindent\textbf{Sampling} As we use normalizing flow to generate $\bm{t}$ and $\bm{q}$, we can draw samples $u$ from the base distribution 
% $p_u(u)$ first and transform them using the bijection $f$ to get samples $f(u)$ in flow's predicted distribution. 
\label{sec:flow-sampling}
\vspace{-3mm}
\subsubsection{End-to-End Training with ContactNet}
\label{sub:subsection_end_to_end}

    %To get physically plausible and high-quality samples from the flow in the GraspGlow, we add loss to the samples drawn from the flow as a constraint. To be specific, in this stage, we first sample rotations using the GraspIPDF and then feed them to the flow as the condition to get translations and joint angles samples. Penetration loss $\mathcal{L}_{\text{pen}}$ and contact map loss $\mathcal{L}_{\text{cmap}}$ are added to make the dexterous hand contact with and grasp the object without penetration. Eventually, the loss in the second stage becomes:

\quad Inspired by~\cite{jiang2021graspTTA}, we use ContactNet to model a mapping from flawed raw grasp pose predictions to ideal contact patterns between the hand and the object. 
The input of the ContactNet is the canonicalized object point cloud $\tilde{X}_0$ and the sampled hand point cloud $\tilde{X}_H$; the output is the contact heat $c_i \in [0, 1]$ predicted at each point $\bm{p}_i\in\tilde{X}_0$. The ground truth of contact heat is given by $ c_i = f(D_i(\tilde{X}_H)) = 2 - 2\cdot (\mathrm{sigmoid}(\beta\,D_i(\tilde{X}_H)))$, where $\beta$ is a coefficient to help map the distance to $[0, 1]$, and $D_i(\tilde{X}_H) = \min\limits_j{\lVert \bm{p}_i - \bm{p}_j \rVert}_2 $, $\bm{p}_j \in \tilde{X}_H$. 
    
Leveraging ContactNet, we construct a self-supervised task to improve the sample quality of GraspGlow by training it end-to-end with RotationNet. To be specific, in this stage, we first sample rotations with GraspIPDF, use them to canonicalize point clouds, then feed the point clouds to GraspGlow to get translations and joint angles samples. Next, ContactNet takes the grasp samples, and outputs ideal contact maps. Here we use four additional loss terms: 1) $\mathcal{L}_{\rm cmap}$: MSE between current and target contact map; 2) $\mathcal{L}_{\rm pen}$: Total penetration from object point cloud to hand mesh calculated using signed squared distance function; 3)$\mathcal{L}_{\rm tpen}$: Total penetration depth from some chosen hand key points to the plane; 4)$\mathcal{L}_{\rm spen}$: Self penetration term inspired by~\cite{zhu2021toward}. Then the joint loss becomes $\mathcal{L}_{\text{joint}}=\mathcal{L}_{\text{NLL}}+\mathcal{L}_{\text{add}}$ where $\mathcal{L}_{\text{add}}$ is defined as:%Finally, a MSE loss $\mathcal L_{\text{cmap} }$ is enforced between the ideal and actual contact maps. We combine $\mathcal L_{\text{cmap}}$ with a penetration loss $\mathcal L_{\text{pen}}$ to form the supervision for the second stage of GraspGlow training. 
%Note that in this stage, ground truth grasp labels are intractable. 
 
\begin{equation}
\mathcal{L}_{\text{add}}=\lambda_{\text{cmap}}\mathcal{L}_{\text{cmap}}+\lambda_{\text{pen}}\mathcal{L}_{\text{pen}}+\lambda_{\text{tpen}}\mathcal{L}_{\text{tpen}}+\lambda_{\text{spen}}\mathcal{L}_{\text{spen}}
\end{equation}

In this stage, we freeze RotationNet as experiments demonstrate that it learns quite well in its own stage.  
% \todo{for implementation details and blablabla please refer to supp sec blablabla, any constants mentioned in main, there must be an explanation in supp}

\vspace{-2mm}
\subsubsection{Test-Time Contact-based Optimization}
\label{sec:tta}
\quad Since the randomness of the flow sometimes leads to small artifacts, the raw outputs of GraspGlow may contain slight penetration and inexact contact. So we use ContactNet to construct a self-supervised optimization task for test-time adaptation to adjust the imperfect grasp as in \cite{jiang2021graspTTA}. 

When GraspGlow predicts a grasp, ContactNet takes the scene and hand point cloud, then outputs a target contact map on the scene point cloud. Next, the grasp pose is optimized for 300 steps to match this contact pattern. The total energy $E_{\mathrm{TTA}}$ consists of the four additional loss term described in Sec.~\ref{sub:subsection_end_to_end}: 
\begin{equation}\small
\lambda^{\rm TTA}_{\rm cmap}E_{\rm cmap}+\lambda^{\rm TTA}_{\rm pen}E_{\rm pen}+\lambda^{\rm TTA}_{\rm tpen}E_{\rm tpen}+\lambda^{\rm TTA}_{\rm spen}E_{\rm spen}
\end{equation}

For network structures, hyperparameters, and other implementation details, please refer to {\color{purple}Sec. B.1.1} of our supp.

% \vspace{-2mm}
\subsection{Goal-Conditioned Dexterous Grasping Policy}
\label{sec:dexterous_grasping_policy}

\begin{figure}
\centering
\includegraphics[width=1\columnwidth]{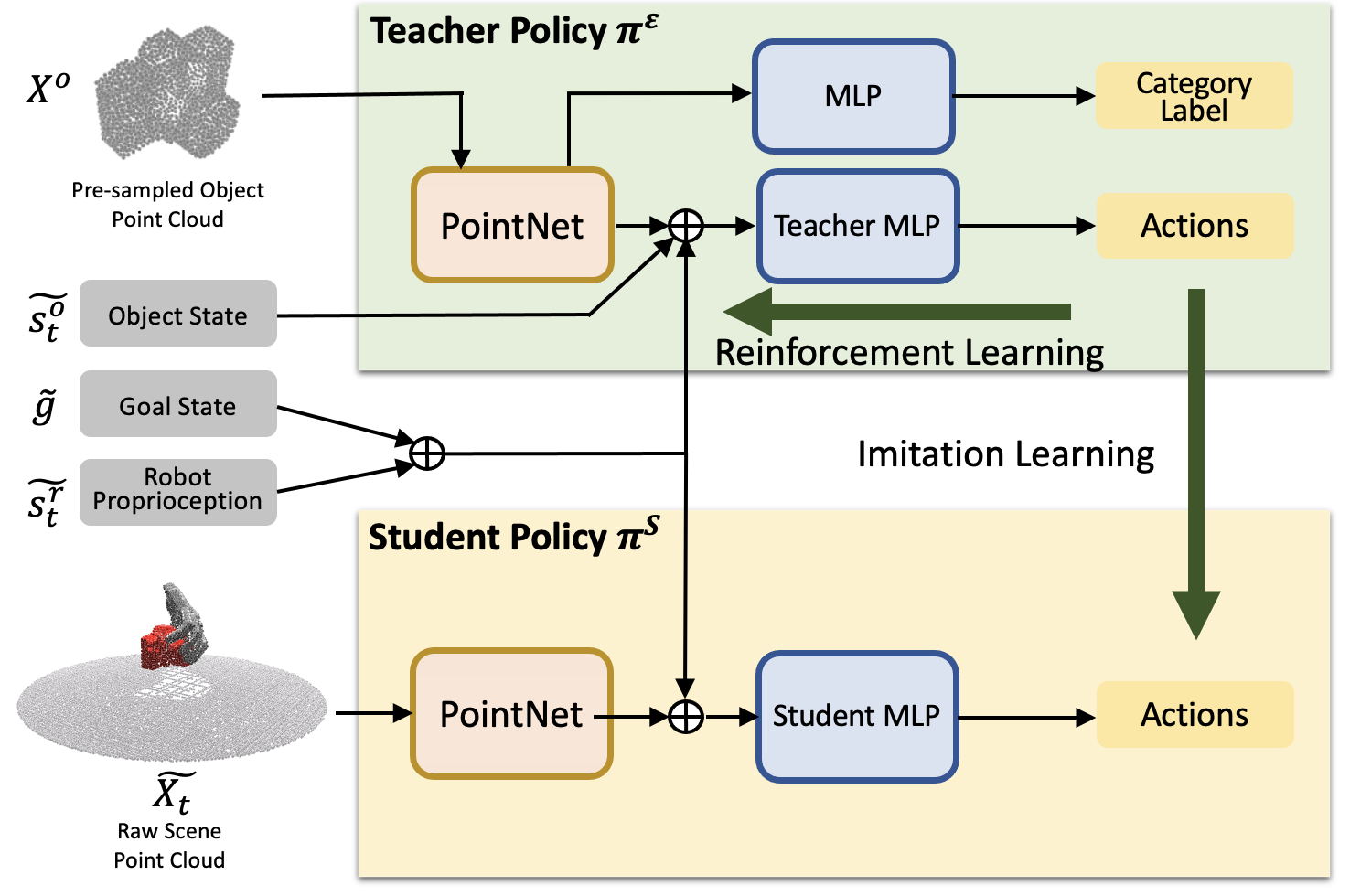}
\caption{\textbf{The goal-conditioned dexterous grasping policy pipeline}. $\widetilde{{\mathcal{S}}^{\mathcal{E}}_t}=(\widetilde{\bm{s}^r_t},\widetilde{\bm{s}^o_t},X^O,\widetilde{g})$ and $\widetilde{{\mathcal{S}}^{\mathcal{S}}_t}=(\widetilde{\bm{s}^r_t},\widetilde{X_t},\widetilde{g})$ denote the input state of the teacher policy and student policy after state canonicalization, respectively;
$\oplus$ denotes concatenation.
}
\label{fig:policyPipeline}
\vspace{-3mm}
\end{figure}

% In this section, we will introduce our \textit{goal-conditioned} grasp policy learning. The task requires the robot hand to grasp and lift the object. 
% Our policy only takes realistic inputs that are available in the real world: the raw scene point cloud, the hand proprioception, and the goal hand grasp label.
%\todo{this information should be in method overview}
% The policy should work across different object categories and even unseen categories.

% Directly training such a vision-based policy using reinforcement learning is challenging\cite{mandikal2021dexvip,mandikal2021graff,mu2021maniskill,shen2022learning}. Inspired by~\cite{chen2021system,chen2020learning}, we use the idea of \textit{teacher-student learning}.
% We first use a popular on-policy RL algorithm, PPO\cite{schulman2017proximal}, to learn an oracle teacher policy that can access the ground-truth states of the environment (\eg object poses, velocities and object full point cloud).
% This information is very useful to the task and available in the simulator but not in the real world. 
% Once the teacher finishes training, we use an imitation learning algorithm, DAgger\cite{ross2011reduction}, to distill this policy to a student policy that can only access the realistic inputs.

%\todo{the header for this subsection is too long, some information should be in related work and intro to motivate our method}
In this section, we will introduce our \textit{goal-conditioned} grasp policy learning. We introduce our proposed state canonicalization, object curriculum learning and other method details for training the teacher policy in Sec.~\ref{sec:teacher}. We then introduce the vision-based student policy training in Sec.~\ref{sec:student}. The teacher policy ${\mathcal{\pi}}^{\mathcal{E}}$ has a state space ${\mathcal{S}}^{\mathcal{E}}_t=(\bm{s}^r_t,\bm{s}^o_t,X^O,\bm{g})$ where $\bm{s}^r_t$ is the robot hand proprioception state, $\bm{s}^o_t$ is the object state, $X^O$ is the pre-sampled object point cloud and $\bm{g}$ is the goal grasp label. The state of the student is defined as ${\mathcal{S}}^{\mathcal{S}}=(\bm{s}^r_t,X_t,\bm{g})$ where $X_t$ is the raw scene point cloud. Details about the state and action space are provided in {\color{purple}Sec. C.1} of our supp.

\vspace{-2mm}
\subsubsection{Learning Teacher Policy}
\label{sec:teacher}
\quad We use a model-free RL framework to learn the oracle teacher policy. The goal of the goal-conditioned RL is to maximize the expected reward $\mathbb{E}\left[  {\textstyle \sum_{t=0}^{T-1}}{\mathcal{\gamma}}^{t}\mathcal{R}(\bm{s}_t,\bm{a}_t,g)\right]$ with ${\mathcal{\pi}}^{\mathcal{E}}$. We use PPO~\cite{schulman2017proximal} for policy updating in our method.
Inspired by ILAD\cite{wu2022learning},
we use a PointNet\cite{qi2016pointnet} to extract the geometry feature of the object. ILAD pre-trains the PointNet using behavior cloning from the motion planning demonstrations and jointly trains the PointNet using behavior cloning from the RL demonstrations during policy learning. 
Although ILAD performs very well in one single category (96\% success rate\cite{wu2022learning}), we find that directly using this method cannot get good results under the setting of \textit{goal-conditioned} and \textit{cross-category}. We then propose several techniques on top of it. 
First, we do state canonicalization according to the initial object pose which improves the sample efficiency of the RL with diverse goal inputs.
Second, we design a novel goal-conditioned function.
We then do 3-stage curriculum policy learning which significantly improves the performance under \textit{cross-category} setting.
Additionally, we find that doing object category classification when joint training the PointNet using behavior cloning which is proposed in \cite{wu2022learning} can also improve the performance under \textit{cross-category} setting.

\label{sec:teacher}
\noindent\textbf{State Canonicalization}
Ideally, this goal-conditioned grasping policy should be $\SO2$ equivariant, that is, when we rotate the whole scene and the goal grasp label with the same angle $\phi$ about the $z$ axis (gravity axis), the grasping trajectory generated by the policy should rotate in the same way.
To ensure this $\SO2$ equivariance, we define a static reference frame (denoted by $\widetilde{\cdot}$ ) according to the initial hand pose: the origin of the reference frame is at the initial hand translation $(0,0,h_0)$ and the Euler angle of the reference frame is $(0, 0, \phi)$ so that the initial Euler angle of the hand in this reference frame is always a fixed value $\widetilde{R_0}=(\frac{\pi}{2}, 0, 0)$.
Before we input the states to the policy, we transfer the states from the world frame to this reference frame: $\widetilde{{\mathcal{S}}^{\mathcal{E}}_t}=(\widetilde{\bm{s}^r_t},\widetilde{\bm{s}^o_t},X^O,\widetilde{g})$. Thus, the system is $\SO2$ equivariant to $\phi$. 
This improves the sample efficiency of the goal-conditioned RL.

\noindent\textbf{Object Curriculum Learning}
Since it's difficult to train the grasping policy \textit{cross-category} due to the topological and geometric variations in different categories, we propose to build an object curriculum learning method to learn ${\mathcal{\pi}}^{\mathcal{E}}$. We find that ${\mathcal{\pi}}^{\mathcal{E}}$ can 
already perform very well when training on one single object but fails when training directly on different category objects simultaneously. We apply curriculum learning techniques. Our curriculum learning technique is constructed as
follows: first train the policy on one single object and then on different objects in one category, several representative categories, and finally on all the categories. We find that this 3-stage curriculum learning significantly boosts the success rates. More details about the ablations on the curriculum stages are in Sec.~\ref{sec:rl_result} and Tab.~\ref{tab:MainAblation}.

\noindent\textbf{Pre-training and Joint Training PointNet with Classification}
ILAD\cite{wu2022learning} proposed an important technique that jointly does geometric representation learning using behavior cloning when doing policy learning. We use this technique in our method and add additional object category classification objectives to update the PointNet.

\noindent\textbf{Goal-conditioned Reward Function}
We are supposed to conquer the dexterous manipulation problem with RL, therefore the reward design is crucial.
Here is our novel goal-conditioned reward function which can guide the robot to grasp and lift the object by the standard of the goal grasp label:
    $
        r = r_{\text{goal}} +  r_{\text{reach}} +  r_{\text{lift}} +  r_{\text{move}}
    $.

The goal reward $r_{\text{goal}}$ punished distance between the current hand configuration and the goal hand configuration. The reaching reward $r_{\text{reach}}$ encourages the robot fingers to reach the object. The lifting reward $r_{\text{lift}}$ encourages the robot hand to lift the object. It's \textbf{non-zero if and only if} the goal grasp label is reached within a threshold. The moving reward $r_{\text{move}}$ encourages the object to reach the target.

\vspace{-3mm}
\subsubsection{Distilling to the Vision-based Student Policy}
\label{sec:student}
\quad We then distill the teacher policy ${\mathcal{\pi}}^{\mathcal{E}}$ into the student policy ${\mathcal{\pi}}^{\mathcal{S}}$ using DAgger\cite{ross2011reduction} which is an imitation method that overcomes the covariate shift problem of behavior cloning. We optimize ${\mathcal{\pi}}^{\mathcal{S}}$ by: ${\mathcal{\pi}}^{\mathcal{S}}= \mathop{\arg\min}\limits_{{\mathcal{\pi}}^{\mathcal{S}}} \lVert  {\mathcal{\pi}}^{\mathcal{E}}({\mathcal{S}}^{\mathcal{E}}_t)-{\mathcal{\pi}}^{\mathcal{S}}({\mathcal{S}}^{\mathcal{S}}_t) \rVert_2$. We also use state canonicalization but this time $\phi$ is the initial Euler angle of the robot hand root around the $z$ axis in the world frame because we don't know the object pose in the student input states. Similarly, we transfer the states from the world frame to this reference frame: $\widetilde{{\mathcal{S}}^{\mathcal{S}}_t}=(\widetilde{\bm{s}^r_t},\widetilde{X_t},\widetilde{g})$.
The pipeline/network architecture is shown in Fig.~\ref{fig:policyPipeline}. For implementation details, please refer to {\color{purple}Sec. B.2.1} of our supp.

\section{Experimentals}
\label{sec:experiments}
\subsection{Data Generation and Statistics}
\label{sec:data_generation}
\noindent We used a similar method from \cite{wang2022dexgraspnet} to synthesize grasps. 

\begin{table*}[!h]\small
    \centering
    \begin{tabular}{@{}l|cccccccccc}
    \toprule
        Method & \multicolumn{2}{c}{seen cat} & \multicolumn{2}{c}{unseen cat} & $\sigma_{\text{$R$}}  \uparrow$ & $\sigma_{\text{$T|R$}}  \uparrow $ & $\sigma_{\text{$\theta|R$}} \uparrow$ &  $\sigma_{\text{keypoints}} \uparrow$ \\
        \cline{2-3} \cline{4-5}
         & $Q_1\uparrow$ & obj. pen.$\downarrow$ & $Q_1\uparrow$ & obj. pen.$\downarrow$ & (degree) & (cm) & (degree) & (cm) \\
        \hline \hline
        GraspTTA\cite{jiang2021graspTTA} (C + T) & 0.0269 & 0.354  & 0.0239 & 0.363 & 4.9  & / & / & 2.909 \\
        DDG~\cite{liu2020deep}  & 0.0357 & 0.319 & 0.0223 & 0.338 &  0.0 & / & / & 0.000 \\
        R + C + T & \underline{0.0362} & 0.251 & \textbf{0.0336} & 0.235 &  \textbf{128.0} & \underline{0.095} & \underline{0.227} & 5.982 \\
        ReLie~\cite{falorsi2019reparameterizing} + T & 0.0190 & 0.219 & 0.0191 & 0.225 & 109.9 & / & / & \textbf{6.698} \\
        ProHMR~\cite{kolotouros2021probabilistic} + T & 0.0210 & \textbf{0.202}  & 0.0221 & \textbf{0.192} & 88.4 & / & / & 5.837 \\
        
        % C & 0.0150 & 0.461 & 0.0026 & 0.356 &  0.0001 & / & / & / \\
        % R + C & 0.0193 & 0.455 & 0.0161 & 0.463 & \textbf{5.294} & 0.0090 &  0.0003 & / \\
        % ReLie~\cite{falorsi2019reparameterizing} & 0.0007 & 0.622 & 0.0004 & 0.552 & 3.742  & / & / & -1.518 \\
        % ProHMR~\cite{kolotouros2021probabilistic} &  &  &  &  & 2.014 & / & / & \todo{-1.709? discuss this} \\
        
        ours (R + GL + T) & \textbf{0.0423} & \underline{0.205} & \underline{0.0322} & \underline{0.220} & \underline{127.6} & \textbf{1.143} & \textbf{5.806} & \underline{6.389} \\
        %\hline 
         %ours (R + GL) & 0.00127 & 0.744  & 0.000 & 0.764  & 128.2 & 1.143 & 5.806 & 7.227 \\
        \bottomrule
         
    \end{tabular}
    \caption{\textbf{Results on grasp goal generation.} R: GraspIPDF, C: CVAE, T: test-time adaptation, GL: GraspGlow, and obj. pen. is the penetration between the hand and the object. }%    Note that the standard deviation of translation and joint angles reported here and in Table 1, the table with the same title in the main paper, is evaluated without test-time adaptation and the same rotations sampled using GraspIPDF are used here to canonicalize the point cloud, and that in Table 1 in the main paper the variance is calculated as $\frac{1}{n}\sum_{i=1}^n\Vert\theta- \overline{\theta}\Vert_2 $, but here the variance is calculated as $\frac{1}{n}\sum_{i=1}^n(\frac{1}{m}\sum_{j=1}^m(\theta_i^j-\overline{\theta^j})^2)$ for clarity.}
    \label{tab:StaticMain}
\end{table*}
% \vspace{-5mm}

\noindent\textbf{Data Generation}\quad First, we randomly select an object from the pool of our training instances and let it fall randomly onto the table from a high place. Next, we randomly initialize a dexterous hand and optimize it into a plausible grasp. The optimization is guided by an energy function proposed by \cite{wang2022dexgraspnet}. We add an energy term on top of this to punish penetration between the hand and the table. Finally, the grasps are filtered by penetration depth and simulation success in Isaac. Please refers to {\color{purple}Sec. A} of our supp.

\noindent\textbf{Statistics}\quad We generated 1.12 million valid grasps for 5519 object instances in 133 categories. These objects are split into three sets: training instances (3251), seen category unseen instances (754), unseen category instances (1514). 
\vspace{-1mm}    
\subsection{Results on Grasp Proposal Generation}
\label{sec:result_ablation}

    \noindent\textbf{Baselines}\quad \cite{jiang2021graspTTA} takes point cloud as input, generates grasps with CVAE, then performs test-time adaptation with ContactNet. Apart from this baseline, we also designed two ablations to verify the two key designs in our pipeline. First, we substituted GraspGlow for the same CVAE from \cite{jiang2021graspTTA} to demonstrate the severe mode collapse of CVAE. Second, we substituted GraspIPDF with ReLie to demonstrate the problem of discontinuity, as described in Sec.~\ref{sec:flow-related-work}. %\todo{add discription to ReLie}

    \noindent\textbf{Metrics}\quad We use some analytical metrics to evaluate quality and diversity. 1) $Q_1$ \cite{ferrari1992planning}. The smallest wrench needed to make a grasp unstable. 2) Object penetration depth(cm). Maximal penetration from object point cloud to hand mesh. 3) $\sigma^2_{\text{$R$/keypoints}}$. Variance of rotation or keypoints. 4) $\sigma^2_{\text{$T/\theta|R$}}$. Variance of translation or joint angles with fixed rotation. 
    
    \noindent\textbf{Ablation 1: Decoupling. }\quad Tab.~\ref{tab:StaticMain} shows that the $Q_1$ of the fourth row and the fifth row is significantly lower than the last row, implying that the normalizing flow on $\SO3 \times \mathbb R^{3+22}$ failed to learn a good distribution. On the other hand, our model can produce grasps with much higher quality. We argue that this improvement is contributed by rotation factorization, which allows us to use IPDF and GLOW to model the distributions on $\SO3$ and $\mathbb R^{22}$ separately. However, as shown in Tab.~\ref{tab:ablation-main}, further decoupling translation and joint angles will result in worse performance as it is less end-to-end. 
    
\begin{figure}[!t]
    \centering
    \includegraphics[width=200pt]{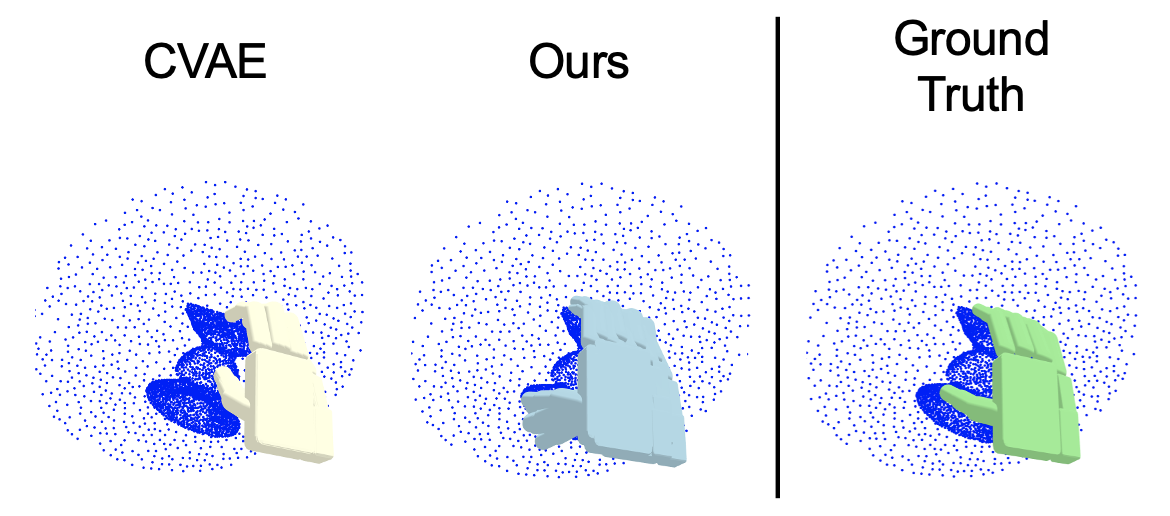}
    \vspace{-2mm}
    \caption{\textbf{Comparison of diversity in grasp translation and articulation given the rotation.} Left: 8 outputs of CVAE (completely collapsed to one pose); Middle: 8 outputs of GraspGLOW; Right: a ground truth grasp.
    %\todo{reorder: CVAE, Ours, (line), GT}~\todo{first line: rotation diversity; second line: given rotation, articulation diversity}
    }
    \label{fig:diversity}
% \vspace{-8mm}
\end{figure}

    \noindent\textbf{Ablation 2: GLOW vs CVAE. }\quad The reason we favored GLOW over CVAE can be interpreted from the last two columns of Tab.~\ref{tab:StaticMain}. If the input object point cloud is fixed, then no matter what the latent $z$ vector is, CVAE will always collapse to a single mode. However, GLOW can propose diverse results. Fig. \ref{fig:diversity} shows a typical case.

    \noindent\textbf{Ablation 3: TTA. }\quad As discussed in Sec.~\ref{sec:tta} and shown in Tab.~\ref{tab:ablation-main}, poses sampled from flow are usually imperfect and need TTA to make them plausible. 

% \begin{table*}
%     \centering
%     \begin{tabular}{@{}l|cccccc}
%     \toprule
%         Method & \multicolumn{2}{c}{seen cat} & \multicolumn{2}{c}{unseen cat} & $\sigma^2_{\text{trans}}  \uparrow $ & $\sigma^2_{\text{joints}} \uparrow$ \\
%         \cline{2-3} \cline{4-5}
%          & $Q_1\uparrow$ & obj. pen.$\downarrow$ & $Q_1\uparrow$ & obj. pen.$\downarrow$ & $(\times 10^{-4})$ \\
%         \hline \hline
%         GraspTTA\cite{jiang2021graspTTA} (C + T) & 0.0269 & 0.354 & 0.0239 & 0.363 & / & / \\
%         R + C + T & 0.0362 & 0.251 & \textbf{0.0336} & 0.235 & 0.0090&0.0003 \\
%         ReLie~\cite{falorsi2019reparameterizing} + T & 0.0190 & 0.219 & 0.0191 & 0.225 & / & / \\
%         ours (R + GL + T) & \textbf{0.0423} & \textbf{0.205} & 0.0322 & \textbf{0.220} & \textbf{1.3063}&\textbf{0.2259}\\
%         \bottomrule
         
%     \end{tabular}
%     \caption{\textbf{Results on grasp goal generation.} R: GraspIPDF, C: CVAE, T: test-time adaptation, GL: GraspGlow, and obj. pen. is the penetration between the hand and the object. GraspTTA and ReLie output rotations which can not be specified so they don't have variations.}
%     \label{tab:StaticMain}
% \end{table*}

\begin{table}\small
    \centering
    \begin{tabular}{@{}c|c|ccc}
    \toprule
        \multicolumn{2}{c|}{Method} & decoup.T & w/o TTA & ours\\
        \hline\hline
        seen   & $Q_1\uparrow$ & 0.0003 & 0.0013 & \textbf{0.0423}\\
        cat.   & pen.$\downarrow$ & 0.862 & 0.744 & \textbf{0.205} \\
        \hline
        unseen & $Q_1\uparrow$ & 0.0061 & 0.0000 & \textbf{0.0322} \\
        cat.   & pen.$\downarrow$ & 0.843 & 0.764 & \textbf{0.220} \\
        \bottomrule
  \end{tabular}
  \caption{\textbf{Ablation study on decoupling translation and joint angles and TTA.} decoup.T: decouple translation, w/o TTA: without TTA and pen. is the penetration between the hand and the object.}
    \label{tab:ablation-main}
\end{table}
% \begin{table}
%   \centering
%   \begin{tabular}{@{}l|cc}
%     \toprule
%     Method & obj./tab. pen.$\downarrow$ & $Q_1\uparrow$  \\
%     % \midrule
%     \hline
%     & \multicolumn{2}{c}{Novel Objects from Known Categories} \\
%     \hline
%     \cite{jiang2021graspTTA} & 0 / 0 & 0\\
%     R + GL + T & 0.298 / 0.682 & \textbf{0.0737} \\
%     R + C + T & \textbf{0.251} / \textbf{0.403} & 0.0427 \\
%     \hline
%      & \multicolumn{2}{c}{Objects from Novel Categories} \\
%     \hline
%     \cite{jiang2021graspTTA} & 0 / 0 & 0\\
%     R + GL + T & 0.299 / 0.795 & \textbf{0.0701} \\
%     R + C + T & \textbf{0.235} / \textbf{0.499} & 0.0427 \\
%     \bottomrule
%   \end{tabular}
%   \caption{Results on static grasp synthesis. R + GL is the pipeline with RotationNet plus GLOW\todo{cite?}; R + C + T is RotationNet, CVAE, and test-time adaptation. sim-succ. is the success rate in simulation. obj./tab. pen. is the penetration on object and table. $E$ is the entropy \todo{equations should be mentioned elsewhere.}. succ-cov. is the success-coverage-curve.}
%   \label{tab:StaticMain}
% \end{table}

% \begin{table}
%   \centering
%   \begin{tabular}{@{}l|cc}
%     \toprule
%     Method & sim-succ. & obj./tab. pen. \\
%     % \midrule
%     \hline
%     \hline
%     GL & 0 & 0 / 0 \\
%     R + GL & 0 & 0 / 0 \\
%     \hline
%     C + T & 0 & 0 / 0 \\
%     R + C + T & 0 & 0 / 0\\
%     \bottomrule
%   \end{tabular}
%   \caption{Ablation study on the RotationNet. R is RotationNet; GL is GLOW\todo{cite}; C + T is CVAE plus test-time adaptation~\cite{jiang2021graspTTA}. }
%   \label{tab:StaticAblation}
% \end{table}

\subsection{Grasp Execution}
\label{sec:rl_result}
%\vspace{-1mm}
\noindent\textbf{Environment Setup and Data}\quad
% We build a table top environment in Isaac Gym for grasp execution. The environment step for each sequence is set to 200 steps. Each environment is randomly initialized with one object and one goal grasp label.
We use a subset of the train split proposed in Sec.~\ref{sec:data_generation} as our training data.
For the state-based policy evaluation, we use the grasp proposals sampled by our grasp proposal generation module.
For the final vision-based policy evaluation, we use both the training data (``GT" in Tab. \ref{tab:MainAblation}) and grasp proposals sampled by our grasp proposal generation module (``pred" in Tab. \ref{tab:MainAblation}) as the goal of our policy to do testing on the train object set and test object set. Details are in {\color{purple}Sec. C.2} of our supp. 
% The test set $1$ contains 420 unseen object instances from the 90 seen categories and the test set $2$ contains 384 unseen object instances from 22 unseen categories.
% During testing, we run the grasp proposal generation module. 
% For each object in our pool of test instances, we sample one rotation from GraspIPDF, canonicalize the scene point cloud, then sample one grasp proposal from GraspGlow. Next, this raw output is adjusted by test-time adaptation using ContactNet. We finally use the output grasp proposal as the goal of our policy to do testing.
% Note that GraspIPDF and GraspGlow output distributions, so it is fully possible to propose more grasp poses and labels, and then select some to execute according to one's need. 

\noindent\textbf{Baselines and Compared Methods}\quad
We adopt PPO\cite{schulman2017proximal} as our RL baseline and DAPG\cite{rajeswaran2017learning} as our Imitation Learning (IL) baseline. We also compared our method with ILAD\cite{wu2022learning} which can reach a very high success rate in one category and can generalize to novel object instances within the same category. We further do ablations on our proposed techniques.
All methods are compared under the same setting of teacher policy. Once we get the teacher policy with the highest success rate, we distill it to the student policy.

\noindent\textbf{Main Results}\quad
The top half of Tab.~\ref{tab:MainRL} shows that our method outperforms baselines by a large margin. The PPO (RL) and DAPG (IL) baseline only achieve an average success rate of 14\% and 13\% on the train set. 
Our teacher method achieves an average success rate of 74\% on the train set, 69\% on the test set, which is about \textbf{49\%} and \textbf{47\%} improvement over ILAD. 
Since our method has a teacher policy with the highest success rate, we distill this policy to a vision-based student policy and the success rate decreases by 6\% and 7\% on the train and test set respectively (the first line of Tab.~\ref{tab:MainAblation}). This indicates that the vision-based, goal-conditional, and cross-category setting is difficult and the student's performance is limited by the teacher's performance.

\noindent\textbf{Ablation Results}\quad %We have done ablation studies on our proposed techniques to show the effectiveness of these techniques.
We evaluate our method without state canonicalization (w/o SC), without  object classification (w/o classification), and without object curriculum learning (w/o OCL). Tab.~\ref{tab:MainRL} shows that each technique yields considerable performance improvement. 
We do more specific ablations on stages of object curriculum learning (OCL).
We stipulate---$\textbf{a}$: train on one object; $\textbf{b}$: train on one category; $\textbf{c}$: train on 3 representative categories; $\textbf{d}$: train on all the categories. We formulate the experiment as follows:
w/o OCL: $\textbf{d}$;
1-stage OCL: $\textbf{a}\rightarrow\textbf{d}$; 
2-stage OCL: $\textbf{a}\rightarrow\textbf{b}\rightarrow\textbf{d}$; 
3-stage OCL: $\textbf{a}\rightarrow\textbf{b}\rightarrow\textbf{c}\rightarrow\textbf{d}$;
As shown in Tab.~\ref{tab:MainRL}, without OCL, the policy seldom succeeds both during training and testing. However, as the total stages of the curriculum increase, the success rate improves significantly.
In Tab.~\ref{tab:MainAblation}, we conduct the robustness test for our vision-based policy by jittering our predicted grasp poses to cause small penetration (a), large penetration (b), and no contact (c) and observe our grasp execution policy is robust to such errors. 

\begin{table}\small
  \centering
  \begin{tabular}{@{}l|ccc}
    \toprule
    Model & Train & \multicolumn{2}{c}{Test}  \\ \cline{3-4} 
    & & \begin{tabular}[c]{@{}c@{}} unseen obj \\ seen cat \end{tabular} & unseen cat \\
    % \midrule
    \hline
    \hline
    MP & 0.12$\pm$0.01 & 0.02$\pm$0.00 & 0.02$\pm$0.01 \\
    PPO\cite{schulman2017proximal} & 0.14$\pm$0.06 & 0.11$\pm$0.04 & 0.09$\pm$0.06 \\
    DAPG\cite{rajeswaran2017learning} & 0.13$\pm$0.05 &  0.13$\pm$0.08 &  0.11$\pm$0.05 \\
    ILAD\cite{wu2022learning} & 0.25$\pm$0.03 & 0.22$\pm$0.04 & 0.20$\pm$0.05 \\
    Ours & \textbf{0.74$\pm$0.07} & \textbf{0.71$\pm$0.05} & \textbf{0.66$\pm$0.06} \\
    \hline
    Ours(w/o SC) & 0.59$\pm$0.06 & 0.54$\pm$0.07 & 0.51$\pm$0.04 \\
    Ours(w/o cls) & 0.65$\pm$0.05 & 0.64$\pm$0.06 & 0.60$\pm$0.07 \\
    Ours(w/o OCL) & 0.31$\pm$0.07 & 0.23$\pm$0.06 & 0.21$\pm$0.04 \\
    Ours(1-stage OCL) & 0.58$\pm$0.07 & 0.55$\pm$0.03 & 0.55$\pm$0.05 \\
    Ours(2-stage OCL) & 0.68$\pm$0.06 & 0.67$\pm$0.07 & 0.62$\pm$0.05 \\
    % Ours (teacher) & \textbf{0.74$\pm$0.07} & \textbf{0.71$\pm$0.05} & \textbf{0.67$\pm$0.09} \\
    % \hline
    % Ours (student) & 0.68$\pm$0.12 & 0.64$\pm$0.09 & 0.60$\pm$0.08 \\
    \bottomrule
  \end{tabular}
  \caption{\textbf{The success rate of state-based policy}. The experiment is evaluated with three different random seeds. ``MP": motion planning;``SC": state canonicalization; ``cls"; joint learning object classification; ``OCL": object curriculum learning.
  }
  \label{tab:MainRL}
\end{table}

\begin{table}\small
  \centering
    \begin{tabular}{@{}l|ccc}
    \toprule
    Penetration (cm) & Train & \multicolumn{2}{c}{Test}  \\ \cline{3-4} 
    & & \begin{tabular}[c]{@{}c@{}} unseen obj \\ seen cat \end{tabular} & unseen cat \\
    % \midrule
    \hline
    \hline
     0.117 (GT) & 0.68$\pm$0.06  &  0.65$\pm$0.05 &  0.63$\pm$0.04 \\
     0.208 (pred) & 0.66$\pm$0.04  &  0.59$\pm$0.04 &  0.58$\pm$0.05 \\
     \hline
     0.512 (a) & 0.63$\pm$0.05  &  0.54$\pm$0.05 &  0.57$\pm$0.04 \\
     1.058 (b) & 0.47$\pm$0.04  &  0.37$\pm$0.05 &  0.39$\pm$0.04 \\
     -0.309 (c) & 0.50$\pm$0.03  &  0.38$\pm$0.02 &  0.35$\pm$0.03\\
    \bottomrule  
  \end{tabular}
  \caption{\textbf{The success rate of our vision-based policy}. We test our trained vision-based policy with jittered goal grasp on the train and test set.}
  \label{tab:MainAblation}
\vspace{-4mm}
\end{table}

\vspace{-1mm}
\subsection{Language-guided Dexterous Grasping}
\begin{figure}
\centering
\includegraphics[width=0.8\columnwidth]{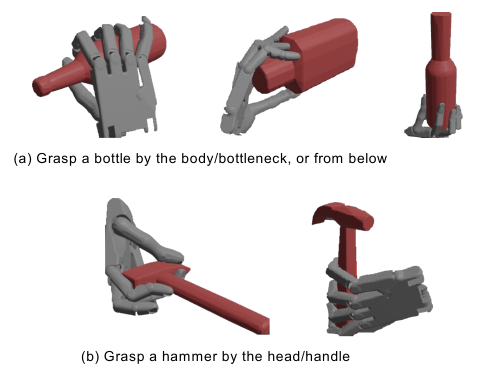}
\caption{\textbf{Qualitative results of language-guided grasp proposal selection.} CLIP can select proposals complying with the language instruction, allowing the goal-conditioned policy to execute potentially functional grasps.}
\label{fig:language}
\vspace{-4mm}
\end{figure}

A natural downstream task for our method is to introduce specific semantic meaning parsing, \eg ``grasping a hammer by the handle''. Thanks to the great diversity of grasp proposals, this can be easily done by adding a filtering module using CLIP\cite{radford2021learning}, a large text-image pre-train model which shows significant generalizability for computing the similarity between text and images. Combined with the goal-conditioned policy network, the robot is endowed with the ability to grasp an object according to human instructions. One of the most promising applications is for the functional grasping of certain tools, \eg bottles and hammers. In detail, we select images rendered from grasp proposals with the highest image-text similarity following the user command (\eg ``A robot hand grasps a hammer by the handle.''). As shown in Fig.~\ref{fig:language}, with only 10 minutes of fine-tuning, the model can achieve around 90\% accuracy in the bottle and hammer categories. This validates the feasibility of generating grasps with specific semantic meanings using our generation pipeline together with the CLIP model.

\vspace{-1mm}
\section{Conclusions and Discussions}

In our work, we propose a novel two-stage pipeline composed of grasp proposal generation and goal-conditioned grasp execution. The whole pipeline for the first time demonstrates universal dexterous grasping over thousand of objects under a realistic robotic setting and thus has the potential to transfer to real-world settings. The limitation is that we only tackle the grasping of rigid objects, in contrast to articulated objects like scissors. Furthermore, functional dexterous grasping still remains a challenging but promising field to explore.
    
    % The results demonstrate that our goal-conditioned dexterous grasping policy has the capability of generalizing well enough to grasp novel objects. Besides, the grasp pose proposal module has also provided a new approach for efficiently synthesizing diverse yet stable dexterous grasps. Also, because the entire pipeline is only based on the object point cloud and the robot's proprioception, this algorithm has the potential to transfer to real world settings.
    
    % We find the divide-and-conquer design paradigm of significant help. When an end-to-end CVAE that predicts the rotation, translation and joint angles altogether fails, we decouple the pipeline into several submodules that serve only one purpose at a time. Similarly, when directly learning an RL algorithm to grasp seems too difficult, we ask it to first learn with the oracle, and then transfer to a more realistic setting.
    
    % Yet there are still some future directions left to explore. Our work only tackles the grasping of rigid objects, in contrast to articulated objects like scissors. If the grasping algorithm takes subsequent manipulation into consideration, it might as well predict a functional grasp and be well informed of the articulation state of the objects in that many tools involve articulation. On one hand, data-driven methods require generating a large dataset with functional grasp labels, which remains an open problem. On the other, training a policy for manipulation without a functional grasp as the guide may not be as trivial as expected. \todo{anything else may be appended}

\clearpage

%%%%%%%%% REFERENCES
{\small
\bibliographystyle{ieee_fullname}
\bibliography{egbib}
}

%%%%%%%%% SUPP

\clearpage
{\large
\textbf{Supplementary Material}
}

\appendix
\renewcommand\thesection{\Alph{section}}

\vspace{4mm}
\noindent\textbf{Abstract}
In this supplementary material, we provide our dataset generation method in Section \ref{sec:dataset-gen}, details about our method, baselines, and implementation in Section \ref{sec:method-detail}, details about the experiments in Section \ref{sec:exp-details}, and more quantitative and qualitative experiment results, in Section \ref{sec:add-result}. For more visualization of our generated dataset and grasping demonstrations, please refer to the supplementary video.

\section{Dataset Generation}
\label{sec:dataset-gen}
In order to train our vision model and RL policy to be universal and diverse in the table-top setting, we need a dexterous grasping dataset that provides numerous object instances and holds diverse grasping labels. Moreover, each grasp should correspond to a physically plausible tabletop scene that is free of any penetration. We synthesized this dataset using a similar method from \cite{wang2022dexgraspnet}.

\noindent\textbf{Object Preperation}\quad Our object dataset is composed of 5519 object instances in 133 categories selected from ShapeNet, \cite{liu2020deep}, and \cite{todorov2012mujoco}. Each object instance is canonicalized into a unit sphere, then re-scaled by each factor in $\{0.06, 0.08, 0.1, 0.12, 0.15\}$. We decompose our meshes into convex pieces using \cite{coacd}. 

\begin{figure}[h!]
    \centering
    \includegraphics[width=230pt]{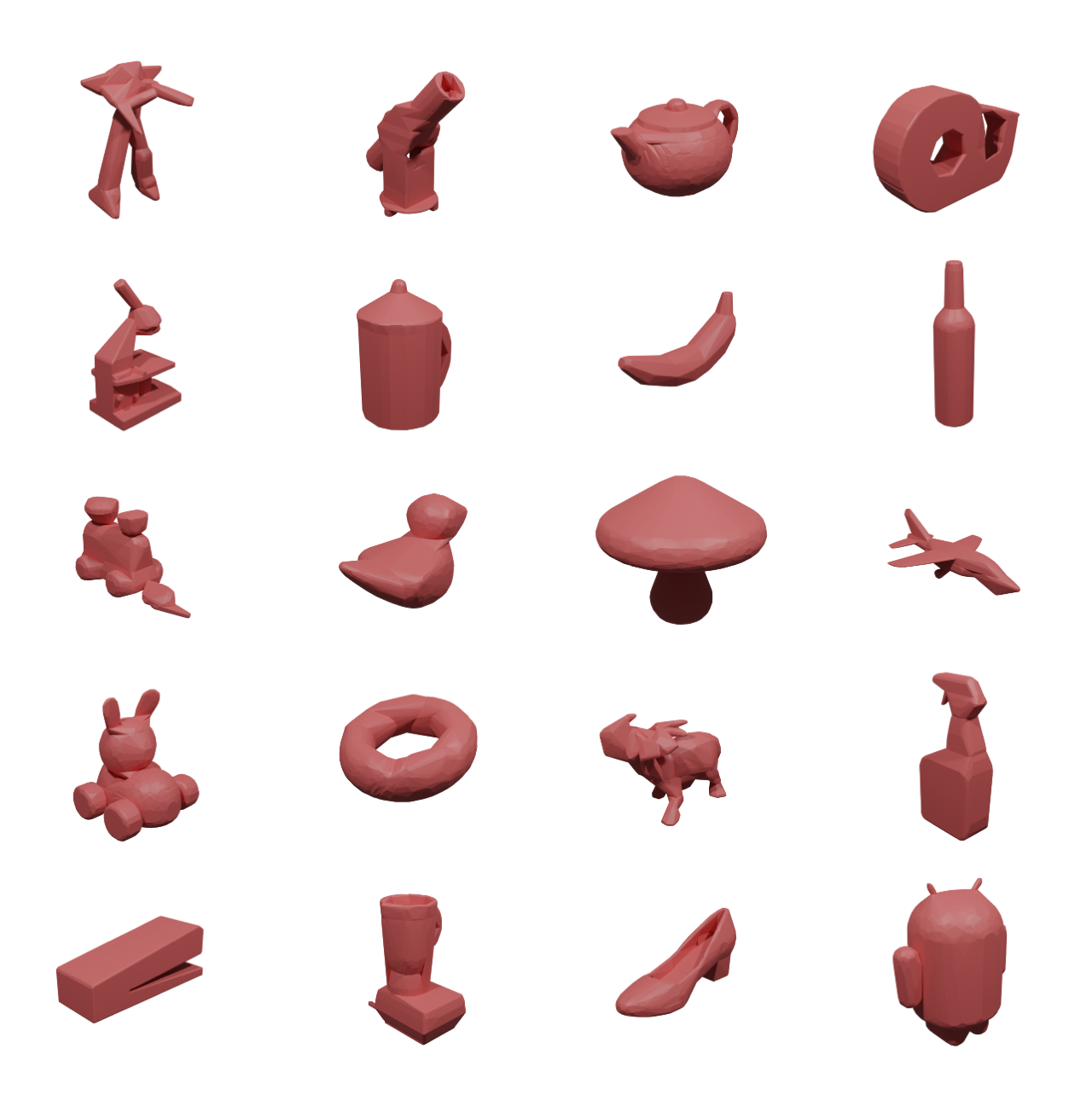}
    \caption{Our object dataset contains more than five thousand objects from various categories. These are the visualization of some decomposed meshes. }
    \label{fig:objects}
\end{figure}

\noindent\textbf{Grasp Generation}\quad Our table-top scene starts with a flat plane which overlaps with the $z=0$ plane of the world reference frame. For each generation environment, we randomly select an object from the pool of our training instances, randomly rotate it, then let it fall onto the plane from a high place. Next, we randomly initialize a dexterous gripper in an area above the object, and let it face the object. Then, the initial gripper pose is optimized into a plausible grasp in a 6000-step optimization process, guided by an energy function. Finally, the object's pose and the gripper's translation, rotation, and joint angles are saved for further validation.

\noindent\textbf{Energy Function}\quad We base our energy function on \cite{tengyuDifferentiableForceClosure}, and modified it to suit the table-top setting. It is composed of the following terms. 1) $E_{\rm fc}$: A differentiable force closure estimator that encourages physical stability; 2) $E_{\rm dis}$: Attraction energy to ensure contact; 3) $E_{\rm pen}$: Repulsion energy to elliminate penetration;  4) $E_{\rm tpen}$: L1 energy that keeps the gripper above the table; 5) $E_{\rm joints}$: Regularization term to enforce joint limits; 6) $E_{\rm spen}$: Self penetration energy inspired by \cite{zhu2021toward}. The total energy is a linear combination of the six terms. 
\begin{equation}
\begin{aligned}  
    E = E_{\text{fc}} + w_{\text{dis}}E_{\text{dis}} + w_{\text{pen}}E_{\text{pen}} +\\ w_{\text{tpen}}E_{\text{tpen}} + w_{\text{joints}}E_{\text{joints}} + w_{\text{spen}}E_{\text{spen}}
\end{aligned}  
\end{equation}

\noindent\textbf{Grasp Validation}\quad We filter our generated dataset by physical stability and penetration depth. A grasp is considered physically stable if it can resist gravity in all 6 axis-aligned directions in the IsaacGym simulator. Moreover, we discard grasps that penetrate the object for more than $1{\rm mm}$. We ran 1000 generations for each object instance, and harvested 1.12 million valid grasps in total. This table-top dataset features stability and diversity, which empowers our models to learn object-agnostic dexterous grasping in a table-top scene. 

\noindent\textbf{Hand SDF Calculation}\quad The signed distance function from the object point cloud to the hand mesh is needed when calculating the penetration energy. However, this forms a batched points-to-mesh distance calculation problem with different meshes, which is hard to compute. So we add some tricks to speed up this calculation. First, we use the collision mesh of the Shadowhand, which is composed of articulation of simple primitives namely boxes and capsules. Second, for a batch of object point clouds, we consider each link of the hand respectively. Third, we use forward kinematics to transform the object point clouds into the link's local reference frame. This operation turns the problem into a points-to-mesh distance calculation with a single mesh. The meshes for boxes are simple, so we use Kaolin \cite{murthy2019kaolin} to compute points-to-mesh distances. As for capsules, the signed distance functions can be defined analytically. Finally, for each point, we take the minimal signed distance across all links to get the signed distance from that point to the complete hand collision mesh.

\section{Method and Implementation Details}

\label{sec:method-detail}

\subsection{Goal Proposal Generation}
\subsubsection{Details about Our Method}
\noindent\textbf{GraspIPDF:}
% \quad Following~\cite{implicitpdf2021}, we implement our GraspIPDF as a function $f: \mathbb{R}^{N\times 3} \times \SO3 \rightarrow \mathbb{R}$. 
\quad The original IPDF~\cite{implicitpdf2021} is a probabilistic model over $\SO3$ that maps a pair of input and rotation, \ie, $(\mathcal{X}, R)$, where the input $\mathcal{X}$ in IPDF is an image, to an unnormalized joint log probability that indicates the likelihood of this pair. Following this work, we implement our GraspIPDF as a function $f(X_0,R): \mathbb{R}^{N\times 3} \times \SO3 \rightarrow \mathbb{R}$. 
We choose PointNet++~\cite{qi2017pointnetplusplus} to extract a global feature from the input point cloud $X_0$, and concatenate it with rotation representation using the same positional encoding in~\cite{implicitpdf2021}. The concatenated feature is processed by an MLP with layer sizes $(256, 256, 256)$ to output $f(X_0, R)$ as formulated in the paper.\\

\noindent\textbf{GraspGlow:}\quad \label{sub:glow-detail} We use Glow~\cite{kingma2018glow} to implement normalizing flow in this part. Glow implements its bijective transformation $f:\mathbb{R}^{3+K} \rightarrow \mathbb{R}^{3+K}$ as the composition of several blocks, where each block consists of three parts: actnorm, $1\times1$ convolution, and affine coupling layer. 

    Our implementation of actnorm and $1 \times 1$ convolution is similar to the implementation in Glow, and the only difference is that our flow is used to transform a single vector of $\mathbb{R}^{3+K}$ instead of an image. Actnorm is a linear function $f_{\rm actnorm}(x)=x/\sigma_\theta+\mu_\theta$ where $\mu_\theta, \sigma_\theta \in \mathbb{R}^{3+K}$ are initialized using the first batch's mean and standard deviation and then optimized with gradient descent like other parameters. $1\times1$ convolution, which can be written as $f_{\rm conv}(x)=Wx$ where $W$ is a $(3+K)\times(3+K)$ invertible matrix, is a linear transformation used as a generalization of a permutation operation. To constraint $W$ to be invertible, $W$ is parametrized with LU decomposition $W=PL(U+S)$ where $P$ is a random orthogonal matrix fixed in the training process, $L$ is a lower triangular matrix with ones on the diagonal, $U$ is an upper triangular matrix with zeros on the diagonal, and $S$ is a diagonal matrix whose diagonal elements are ensured to be positive using $\exp$.
    %, so $s$ and $b$ in actnorm are vectors of length $3+K$, and $P$, $L$, and $U$ in $1\times1$ convolution are matrices of shape $(3+K)\times(3+K)$. For more details, we refer the readers to Glow~\cite{kingma2018glow}. 
    We modify the affine coupling layer in a similar way to ProHMR\cite{kolotouros2021probabilistic}, which can be described as follows:
    \begin{gather*}
        (x_1, x_2) = \operatorname*{split}(x) \\
        (\log{s}, b) = \operatorname*{NN}(x_2, c) \\ 
        s = \exp{(\log{s})} \\
        y_1 = x_1 \\
        y_2 = s \odot x_2 + b \\
        f_{\rm coup}(x) = \operatorname*{concat}(y_1, y_2)
    \end{gather*}
    where $x$ is the input of the transformation, $c\in\mathbf{R}^c$ is the feature extracted by PointNet, $x_1$ is the first half dimensions of $x$ and $x_2$ is the last half dimensions.

    In GraspGlow, we compose 21 blocks described above, and the NNs in each block's coupling layer has 2 residual blocks containing MLPs with two layers, and the number of hidden dimensions is 64. The activation function is ReLU and we use batch normalization in MLPs and the probability of dropout is 0.5. For more details, we refer the reader to ProHMR~\cite{kolotouros2021probabilistic}'s code as we use their implementation of conditional Glow. \\
    % In actnorm, parameters $\mu$ and $\sigma$ are initialized using the first batch's mean and variance and are optimized with other learnable parameters with gradient descent. In $1 \times 1$ convolution, we randomly initialize . See details in Glow~\cite{kingma2018glow}.
    %In the first stage of training, .

\noindent\textbf{ContactNet:}\quad The ContactNet has 2 independent PointNet~\cite{qi2016pointnet} modules respectively for the canonicalized object point cloud $\tilde{X}$ and the hand point cloud $X_H$ sampled from the hand mesh constructed by the forward kinematics. The global feature from the hand is broadcast and concatenated to the per-point feature of the object point cloud. Afterward, the feature goes through an MLP with layer sizes $(1024, 512, 512, 256, 256, 128, 128, 10)$ to output the 10-bin-discretized contact map per-point prediction. The discretization is found to greatly boost the robustness of our ContactNet.

\subsubsection{Details about Baselines}

\noindent\textbf{GraspTTA:}\quad 
GraspTTA~\cite{jiang2021graspTTA} proposed a two-stage framework to synthesize grasps for MANO~\cite{MANO}. They designed a CVAE and a ContactNet to perform the tasks. During training, the CVAE learns to reconstruct the grasp dataset, and the ContactNet learns a mapping from the object and hand point cloud to the object's contact map. During testing, in the first stage, the CVAE takes the object point cloud's feature as a condition, samples a latent vector from the Gaussian distribution, and outputs the hand rotation, translation, and parameters. They use these to reconstruct the hand mesh. In the second stage, the ContactNet takes the object and hand point cloud, and predicts a target contact map on the objects. The consistency energy is defined as the MSE between the actual contact map and the target contact map predicted by the contact net. Using this energy, the hand is optimized toward the target in a test-time adaptation process. Note that in their work, the target contact map is recalculated in every optimization step. We also use test-time adaptation in our pipeline, but only calculate the target contact map in the first iteration to save time. 
 \\

\noindent\textbf{DDG:}\quad
Deep Differentiable Grasp (DDG) \cite{liu2020deep} takes 5 depth images of the object and regresses the translation, rotation, and joint angles of the ShadowHand. The learning process is divided into two stages. In the first stage, only a min-of-N loss is used for grasp pose regression. In the second stage, other loss functions are added to encourage contact, avoid penetration, and improve grasp quality. It is important to note that, the original method doesn't take the table as input. They assume that all depth images are taken in the object reference frame, and for each set of depth images, 100 ground-truth grasps are required to define the min-of-N regression loss. However, if we change this setting and take the depth images in the table reference frame, then only 15 ground-truth grasps are available for each set of images on average. This is because when we synthesized data for each object, the table planes are randomly chosen in each generation process. So in this experiment, we preserved their original settings without the table. In evaluation, we don't filter grasps that have large table penetration depth for this method, which makes the problem easier, and prove that our method still out-perform theirs. \\

\label{subsubsec:rotation-flow-baseline}

\noindent\textbf{ReLie:}\quad ReLie~\cite{falorsi2019reparameterizing} proposes a general way to perform normalizing flow on Lie groups using Lie algebra, and in $\SO3$ this equals to using the axis-angle representation $\textbf{v}=\theta\textbf{n}$ where $\theta$ is the angle of the rotation and that $\textbf{n}$ is a unit vector representing the axis of rotation. In their implementation, the normalizing flow is performed on $\mathbb{R}^3$ and then the samples are transformed using $\operatorname{tanh}(\cdot)$ and multiplied by $r$, so that the length of the sampled vector is less than $r$, and at last, the transformed samples are converted to rotation using the exponential map. Note that for surjectivity, they sacrifice invertibility and set $r$ to $1.6\pi$. In our experiment, we add those three dimensions to GraspGlow so that the hand root rotation, translation, and joint angles can be sampled jointly. The problem with this method is that it suffers from discontinuity of the axis angle representation and that the lack of bijectivity is also harmful to the learning process.  \\

\noindent\textbf{ProHMR:}\quad  ProHMR~\cite{kolotouros2021probabilistic} proposes to use a 6D representation of $\SO3$, which is the first two columns of the rotation matrix, to avoid discontinuity. In their method, normalizing flow is performed on $\mathbb{R}^6$ and the samples are projected to the manifold of $\SO3$ afterward. In our experiment, we add those six dimensions to GraspGlow similar to the baseline of ReLie, and to make the samples close to the $\SO3$ manifold, we also add the orthogonal loss following ProHMR. The problem with this method is that the projection is an infinite-to-one mapping so that the probability of a specific rotation is intractable, so theoretically the normalizing flow can place infinite probability to the $\SO3$ manifold without learning distribution on $\SO3$ to get infinitely low NLL. \\

\subsection{Goal-conditioned Dexterous Grasping Policy}
\subsubsection{Details about Our Method}
\quad As we introduced in Sec.~3.3.2, we use PPO to update the teacher policy. We also adopt the technique in ILAD\cite{wu2022learning} which jointly learns object geometric representation by updating the PointNet using behavior cloning objective during RL policy training. We further propose three important techniques including state canonicalization, object curriculum learning, and joint training object classification to update the PointNet in our network. \\

\noindent\textbf{Reward Function:}
\quad To ensure proper interaction between the robot hand and the object and encourage the robot to grasp the object according to the input grasping goal pose, we define a novel goal-conditioned reward function. Since we aim to solve the dexterous manipulation problem with pure RL, the reward design is crucial. Note that all the $\omega_{**}$ here are hyper-parameters.

   The goal pose reward $r_{\text{goal}}$ encourages the robot hand to reach the input grasping goal pose. It measures the weighted sum of distances between current robot joint angles $q_j$ and goal joint angles $q_j^g$, and the distance between hand root pose $(t_h^{obj}, R_h^{obj})$ in the object reference frame and the goal hand root pose $(t_h^{g}, R_h^{g})$ :
   \begin{equation}
   r_{\text{goal}} = - \omega_{g, q}\sum_{j=1}^{J}{ \lvert q_j - q_j^g \rvert} 
   -\omega_{g, t}\lVert t_h^{obj} - t_h^g \rVert_2
   -\omega_{g, R}L_{rot}
   \end{equation}
   
   $M_{R_h^{obj}}$ and $M_{R_h^g}$ are matrices of the object relative hand rotation and goal hand rotation. The $L_{rot}$ here stands for the axis angle from goal hand rotation to object relative hand rotation, and is formulated as follows: 
   
   \begin{equation}
   L_{rot} = \text{acos}(0.5(\text{trace}(M_{R_h^{obj}} M_{R_h^g}^\top)-1))
   \end{equation}

   The reaching reward $r_{\text{reach}}$ encourages the robot fingers to reach the object. Here, $\textbf{x}_{\text{finger}}$ and $\textbf{x}_{\text{obj}}$ denote the position of each finger and object:
   \begin{equation}
   r_{\text{reach}} = - \omega_{r} \sum{\lVert  \textbf{x}_{\text{finger}}-\textbf{x}_{\text{obj}} \rVert_2 }
   \end{equation}

   The lifting reward $r_{\text{lift}}$ encourages the robot hand to lift the object when the fingers are close enough to the object and the robot hand is considered to reach the target goal grasp pose. $f$ is a flag to judge whether the robot reaches the lifting condition:
   $f=\textbf{Is} (\sum_{j=1}^{J}{\omega_{g,j} \lVert \textbf{x}_{j}^{obj}-\textbf{x}_{g,j}^{obj} \rVert_2}<\lambda_{f_1})+\textbf{Is} ( \sum{\lVert  \textbf{x}_{\text{finger}}-\textbf{x}_{\text{obj}} \rVert_2 }<\lambda_{f_2}) + \textbf{Is}({d}_{\text{obj}}>\lambda_{0})$. Here, ${d}_{\text{obj}} = \lVert  \textbf{x}_{\text{obj}}-\textbf{x}_{\text{target}} \rVert_2$, where $\textbf{x}_{\text{obj}}$ and $\textbf{x}_{\text{target}}$ are object position and target position. $a_z$ is the scaled force applied to the hand root along the z-axis ($\omega_{l}>0$).
   \begin{equation}
       r_{\text{lift}} = 
       \begin{cases}
       \omega_{l}*(1+a_z)  & \text{ if } f=3 \\
       0 & \text{ otherwise } 
       \end{cases}
   \end{equation}
    
   The moving reward $r_{\text{move}}$ encourages the object to reach the target and it will give a bonus term when the object is lifted very close to the target:
   \begin{equation}
       r_{\text{move}} = 
       \begin{cases}
       -\omega_{m}{d}_{\text{obj}} + \frac{1}{1+\omega_{b}{d}_{obj}}  & \text{ if } {d}_{\text{obj}}<\lambda_{0} \\
       -\omega_{m}{d}_{\text{obj}} & \text{ otherwise } 
       \end{cases}
   \end{equation}
   Finally, we add each component and formulate our reward function as follows:
   \begin{equation}
   r = r_{\text{goal}} +  r_{\text{reach}} +  r_{\text{lift}} +  r_{\text{move}}
   \end{equation} \\

\noindent\textbf{Details of Object Curriculum Learning:} 
\quad For OCL (Object Curriculum Learning) experiments in Sec.~4.3 in the main paper, here is the detail of the curriculum.

For 1-stage OCL we randomly choose three different categories' objects and do three experiments. Each time first we train on one object and then train on all the categories.

For 2-stage OCL we randomly choose three categories and do three experiments. Each time first we train on one object from each category, then train on the category where the object is from, and last train on all the categories.

For 3-stage OCL, we do two experiments. The 3 representative categories we use in the first experiment are (toy car, bottle, and camera). The 3 representative categories we use in the second experiment are the (light bulb, cereal box, and toy airplane). We first train on one object, then train on the category of the object, then train on 3 representative categories, and last train on all the categories. \\

\noindent\textbf{Network Architecture:}
\quad The MLP in teacher policy ${\mathcal{\pi}}_{\mathcal{E}}$ and student policy  ${\mathcal{\pi}}_{\mathcal{S}}$ consists
of 4 hidden layers (1024, 1024, 512, 512). The network structure of the PointNet in both the ${\mathcal{\pi}}_{\mathcal{E}}$ and ${\mathcal{\pi}}_{\mathcal{S}}$ is (1024, 512, 64). We use the exponential linear unit (ELU)\cite{clevert2015fast} as the activation function.

\subsubsection{Details about Baselines}
\label{sec:details_baselines}

\noindent\textbf{MP (Motion Planning)}
\quad We use cross-entropy method (CEM)\cite{rubinstein1999cross} for motion planning given  target hand joint positions $j^g$ computed from the target goal hand grasp label $\bm{g}$ using forward kinematics. The goal is to find
a robot hand action sequence $a_1, ..., a_K$ which generates a robot hand joint position sequence $\bm{j}^r_0, ..., \bm{j}^r_K$ and the last robot hand joint positions $\bm{j}^r_K$ reaches $\bm{j}^g$.
Followed by \cite{wu2022learning}, the objective of the motion planning is $\text{min}_{a_1,...a_K} \lVert  \bm{j}_K^r-\bm{j}^g \rVert^2 + \lambda \lVert  \textbf{x}_{\text{obj}}^1-\textbf{x}_{\text{obj}}^K \rVert^2$, where $\textbf{x}_{\text{obj}}^1$ and $\textbf{x}_{\text{obj}}^K$ are object poses at time step 1 and $K$. This objective function encourages the robot hand to reach the goal hand grasp label as well as prevents the object from moving during the process. We use model predictive control (MPC) to execute the planned trajectories sampled from CEM process until the objective is below a threshold $\delta$ .
Once the process ends, we lift the robot's hand to see whether the object falls down. \\

\noindent\textbf{PPO}
\quad PPO~\cite{schulman2017proximal} is a popular model-free on-policy RL method. We use PPO together with our designed goal-conditioned reward function as our RL baseline. \\

\noindent\textbf{DAPG}
\quad Demo Augmented Policy Gradient (DAPG)~\cite{rajeswaran2017learning} is a popular imitation learning (IL) method that leverages expert demonstrations to reduce sample complexity. Followed by ILAD\cite{wu2022learning}, we use motion planning to generate demonstrations from our goal grasp label dataset. We use our designed goal-conditioned reward function in this method. \\

\noindent\textbf{ILAD}
\quad  ILAD~\cite{wu2022learning} is an imitation learning method that improves the generalizability of DAPG. ILAD proposes a novel imitation learning objective on top of DAPG and it jointly learns the geometric representation of the object using behavior cloning from the generated demonstrations during policy learning.
For this method, we use the same generated demonstrations as in DAPG and use our designed goal-conditioned reward function. \\

\noindent\textbf{IBS-Grasp}
\quad IBS-Grasp~\cite{she2022learning} propose an effective representation of the grasping state called Interaction Bisector Surface (IBS) characterizing the spatial interaction between the robot hand and the object. 
The IBS representation, together with a novel vector-based reward and an effective training strategy, facilitates learning a strong control model of dexterous grasping with good sample efficiency and cross-category generalizability. 
It uses SAC\cite{haarnoja2018soft}to train the policy. Note that we evaluate the baseline using the official code which uses the Pybullet simulator because it's hard to do the proposed fast IBS approximation in Isaac gym. Additionally, it cannot train under the goal-conditioned setting using the proposed reward function. \\

\section{Experiment Details}
\label{sec:exp-details}
\subsection{Environment Setup}

%\subsubsection{Dataset}

\noindent\textbf{State Definition}\quad
\begin{table}
    \centering
    \begin{tabular}{@{}l|cc}
    \toprule
        Parameters & Description \\ 
        \hline \hline
        $\bm{q} \in \mathbb{R}^{18}$ & joint positions \\
        $\bm{\dot{q}} \in \mathbb{R}^{18}$  & joint velocities \\
        $\bm{\tau}_{\text{dof}} \in \mathbb{R}^{24}$  & dof force  \\
        $x_{\text{finger}} \in \mathbb{R}^{3\times5}$                          
          & fingertip position              \\
        $\alpha_{\text{finger}} \in \mathbb{R}^{4\times5}$                        & fingertip orientation           \\
        $\dot{x}_{\text{finger}} \in \mathbb{R}^{3\times5}$                    
          & fingertip linear velocities     \\
        $\omega_{\text{finger}} \in \mathbb{R}^{3\times5}$  
          & fingertip angular velocities    \\
        $F_{\text{finger}} \in \mathbb{R}^{3\times5}$  & fingertip force   \\
        $\tau_{\text{finger}} \in \mathbb{R}^{3\times5}$ & fingertip torque  \\
        $\bm{t} \in \mathbb{R}^3$  & hand root global transition          \\
        $R \in \mathbb{R}^{3\times3}$  & hand root global orientation         \\
        $\bm{a} \in \mathbb{R}^{24}$   & action \\
        \bottomrule
    \end{tabular}
    \caption{Robot state definition.}
    \label{tab:robot_state}
\end{table}
The full state of the teacher policy ${\mathcal{S}}^{\mathcal{E}}=(\bm{s}^r_t,\bm{s}^o_t,X^O,\bm{g})$.
The full state of the student policy
${\mathcal{S}}^{\mathcal{S}}=(\bm{s}^r_t,X_t,\bm{g})$.
The robot state $\bm{s}_r$ is detailed in Tab.~\ref{tab:robot_state} and the object oracle state $\bm{s}_o$ includes the object pose, linear velocity, and angular velocity.
For the pre-sampled object point cloud $X^O$, we sample 2048 points from the object mesh. For the scene point cloud $X_S$, we only sample 1024 points from the object and the hand to speed up the training. \\

\noindent\textbf{Action Space}
The action space is the motor command of 24 actuators on the robotic hand. The first 6 motors control the global position and orientation of the robotic hand and the rest 18 motors control the fingers of the hand. We normalize the action range to (-1,1) based on actuator specification.\\

\begin{figure}[!t]
    \centering
    \includegraphics[width=0.6\linewidth]{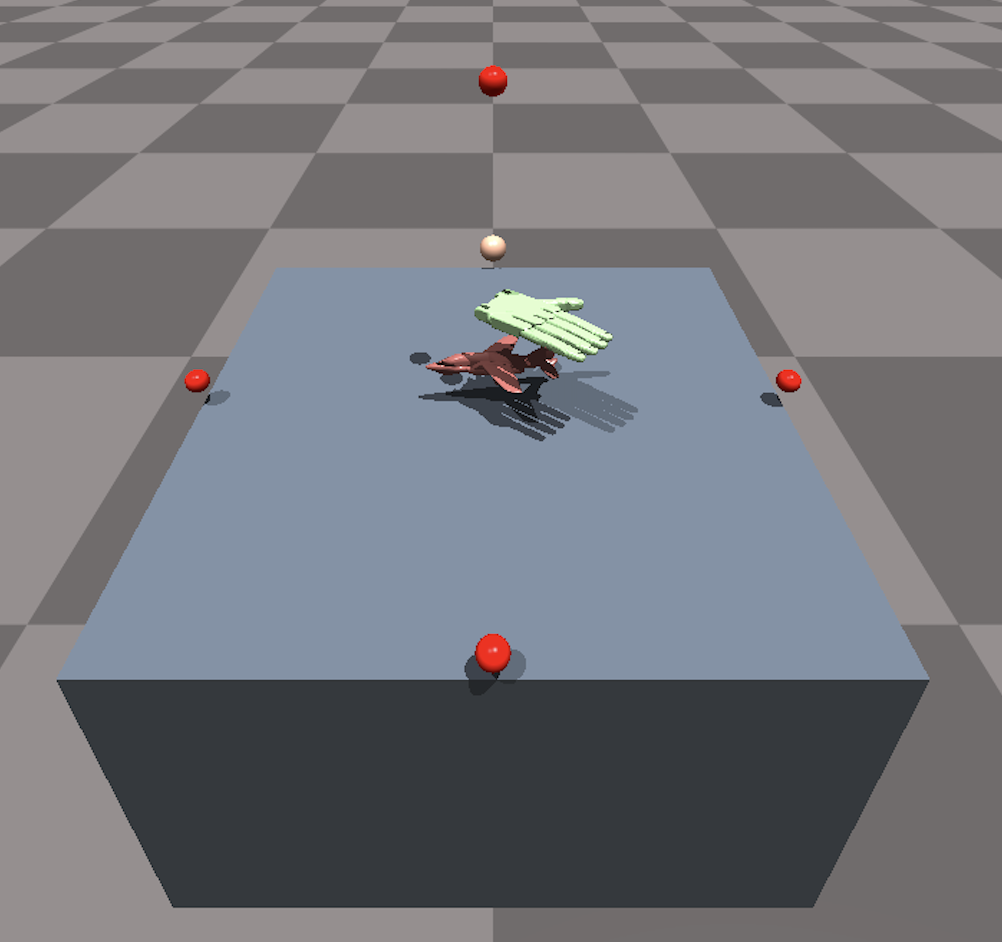}
    \caption{\textbf{Camera positions}}
    \label{fig:camera}
\end{figure}

\noindent\textbf{Camera Setup}\quad
We placed five RGBD cameras, four around the table and one above the table, as shown in Fig.\ref{fig:camera}. The origin of the system is the center of the table. The positions of the five cameras are: ([0.5, 0, 0.05], [-0.5, 0, 0.05], [0, 0.5, 0.05], [0, -0.05, 0.05], [0, 0, 0.55]) and all their focus point is [0, 0, 0.05].

%\subsection{Baselines}
%Here we provide an extended description of the baselines.

%(1) For the goal proposal generation part, we introduce three baselines to validate the design of our method.

%(2) For learning the goal-conditioned dexterous grasping teacher policy:

%\subsection{Experiment Setup}
%\subsubsection{Network Architecture}
%\label{sub:graspglow-detail}
%\quad(1) For the goal proposal generation part:

%Backbone:

%GraspGlow:\todo{to lhr: write this}
%    Backbone  \\

\subsection{Training Details}
(1) 
For the goal proposal generation part,
we use Adam optimizer to train GraspIPDF, with a learning rate of $10^{-3}$ and a batch size of 16. The loss Curve converges in 24 hours. 
In the training process of GraspGlow, we use Adam as our optimizer and the learning rate is $10^{-3}$. The experiment is done on an NVIDIA RTX A5000, and the batch size is set to 64 in the first stage and 32 in the second stage with 8 samples for each object. The training process consists of 160k iterations in the first stage and 8k iterations in the second stage, and it needs one day in total. ContactNet also uses Adam, with a learning rate of $10^{-3}$ and a batch size of 128, and the normalization factor $\beta$ is set to 60. This module needs 8 hours of training. The hyperparameters for end-to-end training and test-time adaptation are in Tab.~\ref{tab:SuppParamVision}. 

(2) For the goal-conditioned dexterous grasping policy part, we use PPO\cite{schulman2017proximal} to learn ${\mathcal{\pi}}_{\mathcal{E}}$ and then distill to ${\mathcal{\pi}}_{\mathcal{S}}$ using DAgger\cite{ross2011reduction}. Since some of the object meshes in the dataset are too large for dexterous grasping, we filter out some large-scale object instances. In the end, we obtain a train set of 3200
object instances and a test set of 241 object instances for the grasping execution experiments. The hyperparameters for the experiments are in Tab.~\ref{tab:SuppParam}.

\begin{table}
    \centering
    \begin{tabular}{l|c}
    \toprule
        Hyperparameter & Value \\
        \hline\hline
        $\lambda_{\mathrm{cmap}}$ & 0.02 \\       
        $\lambda_{\mathrm{pen}}$ & 500 \\
        $\lambda_{\mathrm{tpen}}$ & 50 \\
        $\lambda_{\mathrm{spen}}$ & 10 \\
        \hline
        $\lambda^{\rm TTA}_{\mathrm{cmap}}$ & 0.07\\
        $\lambda^{\rm TTA}_{\mathrm{pen}}$ & 10000\\
        $\lambda^{\rm TTA}_{\mathrm{tpen}}$ & 1000\\
        $\lambda^{\rm TTA}_{\mathrm{spen}}$ & 10 \\
        step size & 0.001
    \end{tabular}
    \caption{\textbf{Hyperparameters for end-to-end training (upper half) and test-time adaptation (lower half).} }
    \label{tab:SuppParamVision}
\end{table}

\begin{table}
    \centering
    \begin{tabular}{@{}l|cc}
    \toprule
        Hyperparameter & Value  \\
        \hline \hline
        Num. envs (Isaac Gym) & 1024 \\
        Env spacing (Isaac Gym) & 1.5\\
        Num. rollout steps per policy update & 8 \\
        Num. batches per agent & 4 \\
        Num. learning epochs & 5 \\
        Episode length & 200 \\
        Discount factor & 0.96 \\
        GAE parameter & 0.95 \\
        Entropy coeff. & 0.0 \\
        PPO clip range & 0.2 \\
        Learning rate & 0.0003 \\
        Value loss coeff. & 1.0 \\
        Max gradient norm & 1.0 \\
        Initial noise std. & 0.8 \\
        Desired KL & 0.16 \\
        Clip observations & 5.0 \\
        Clip actions & 1.0 \\
        $\omega_{g, q}$ & 0.1 \\
        $\omega_{g, t}$ & 0.6 \\
        $\omega_{g, R}$ & 0.1 \\
        $\omega_{r}$ & 0.5 \\
        $\omega_{l}$ & 0.1 \\
        $\omega_{m}$ & 2 \\
        $\omega_{b}$ & 10 \\
        
        \bottomrule
         
    \end{tabular}
    \caption{\textbf{Hyperparameters for grasping policy.} }
    \label{tab:SuppParam}
\end{table}

\subsection{Metric Details}
\label{sec:metric_details}

\begin{table*}[!h]
    \centering
    \begin{tabular}{@{}l|cccccc|ccc}
    \toprule
        Method & \multicolumn{6}{c|}{goal-conditioned} & \multicolumn{3}{c}{non-goal-conditioned}  \\
        \cline{2-10}
         & \multicolumn{2}{c}{Train} & \multicolumn{4}{c|}{Test} & \multicolumn{1}{c}{Train} & \multicolumn{2}{c}{Test} \\
         \cline{4-7} \cline{9-10}
         & \multicolumn{2}{c}{} & \multicolumn{2}{c}{\begin{tabular}[c]{@{}c@{}} unseen obj \\ seen cat\end{tabular}} & \multicolumn{2}{c|}{unseen cat} & \multicolumn{1}{c}{} & \multicolumn{1}{c}{\begin{tabular}[c]{@{}c@{}} unseen obj \\ seen cat\end{tabular}} & \multicolumn{1}{c}{unseen cat} \\
         \cline{2-7} \cline{8-10}
         & succ$\uparrow$ & MPE (cm)$\downarrow$ & succ$\uparrow$ & MPE (cm)$\downarrow$ & succ$\uparrow$ & MPE (cm)$\downarrow$  & succ$\uparrow$ & succ$\uparrow$ & succ$\uparrow$ \\
        \hline \hline
        % MP & 0.00$\pm$0.00 & 0.00$\pm$0.00 & 0.00$\pm$0.00 & 0.00$\pm$0.00 & 0.00$\pm$0.00 & 0.00$\pm$0.00 & 0.00$\pm$0.00 & 0.00$\pm$0.00 & 0.00$\pm$0.00 & 0.00$\pm$0.00 & 0.00$\pm$0.00 & 0.00$\pm$0.00 \\
        MP & 0.12 & \textbf{1.2} & 0.02 & \textbf{1.8} & 0.02  & \textbf{1.8} & / & / & / \\
        PPO\cite{schulman2017proximal} & 0.14 & 4.4 & 0.11 &  4.9 & 0.09 & 5.8 & 0.24 & 0.21 & 0.17  \\
        DAPG\cite{rajeswaran2017learning} & 0.13 & 8.0 & 0.13 & 7.4 & 0.11 & 9.1 & 0.21 & 0.15 & 0.10 \\
        ILAD\cite{wu2022learning} & $\underline{\text{0.25}}$ & 5.1 & $\underline{\text{0.22}}$ & 5.3 & $\underline{\text{0.20}}$ & 5.6 & 0.32 & 0.26 & 0.23 \\
        IBS-Grasp\cite{she2022learning} & / & / & / & / & / & / & $\underline{\text{0.57}}$ & $\underline{\text{0.54}}$ & $\underline{\text{0.54}}$ \\
        Ours (teacher) & \textbf{0.74} & $\underline{\text{3.5}}$ & \textbf{0.71} & $\underline{\text{3.9}}$ &\textbf{0.67} & $\underline{\text{4.5}}$ & \textbf{0.79} & \textbf{0.74} & \textbf{0.71} \\
        \hline \hline
        Ours (student) & 0.68 & 3.8 & 0.64 & 4.3 & 0.60 & 4.7 & 0.74 & 0.69 & 0.65    \\ 
        \bottomrule
         
    \end{tabular}
    \caption{\textbf{Results on dexterous grasping policy.} We use bold to indicate the best metric and underline to indicate the second-best metric. Note that for MP (Motion planning), ``train" means optimizing on our synthetic ground truth grasp dataset and ``test" means optimizing on the predicted grasp from our vision pipeline.}
    \label{tab:SuppStaticPolicy}
\end{table*}

\subsubsection{Metrics for Goal Proposal Generation}
\label{sec:metric_details_goal}

For the goal proposal generation part, we introduce seven metrics to evaluate grasp quality and two metrics to measure the diversity of our generated results. \\

\noindent\textbf{Mean $Q_1$}\quad $Q_1$~\cite{ferrari1992planning} is defined as the minimal wrench needed to make a grasp unstable. This metric is only well defined when the grasp has exact contact and doesn't penetrate the object, which is impossible for vision-based methods. So we relaxed the contact distance to $1\text{cm}$. Moreover, if a grasp penetrates the table for more than $1\text{cm}$, or has an object penetration depth larger than $5\text{mm}$, then this grasp will be considered invalid, so its $Q_1$ will be manually set to zero. \\

\noindent\textbf{Object Penetration Depth}\quad 
We define object penetration depth as the maximal penetration from the object's point cloud to the hand mesh. This is calculated using the tricks we introduced in Sec.~\ref{sec:dataset-gen}. \\

\noindent\textbf{Rotation Standard Deviation}\quad This metric evaluates the standard deviation of rotation by first calculating the chordal L2 mean of rotation samples ($\underset{R}{\operatorname*{argmin}} \sum_{i=1}^n(||R-R_i||_F^2$), and then calculate the standard deviation between the rotation samples and the mean. This metric is used to show the diversity of rotation samples from GraspIPDF. \\

\noindent\textbf{Translation and Joint Angles Standard Deviation (conditional)}\quad These two metrics evaluate the standard deviations of translation and joint angles, given a sampled rotation from GraspIPDF, and are used to show the diversity of translation and joint angles in the grasp samples from GraspGlow.\\

\noindent\textbf{Keypoint Standard Deviation}\quad
This metric evaluates the average standard deviation of 15 joint (keypoint) positions of the robotic hand. This metric is used to show the diversity of our grasp proposals generated by the whole grasp proposal generation pipeline.\\

\noindent\textbf{Log-likelihood}\quad We evaluate the log-likelihood of the ground truth grasping rotation, translation, and joint angles predicted by the model, and this metric is used to show how well the model fits the ground truth distribution. Note that for our model, we calculate $p(R, t, \theta | X)$ as $p(R|X)\cdot p(t, \theta | R^{-1}X)$. \\

%\noindent\textbf{Static Grasping Success}\quad
%We evaluate the success rate of the predicted grasp pose using the simulator. 

\subsubsection{Metrics for Goal-conditioned Grasp Execution}
For the goal-conditioned dexterous grasping policy part, we introduce metrics to measure the success rate of grasping, as well as how strictly our policy follows the specified grasping goal. \\

\noindent\textbf{Simulation Success Rate}\quad We define the success rate as the primary measure of the grasping policy. The target position of the object is 0.3m above its initial position. The task is considered successful if the position difference between the object and the target is smaller than 0.05m at the final step of one sequence. \\

% \noindent\textbf{MPE}\quad This metric is used for measure the mean position error between the hand root position of the exact grasping pose $\bm{t}_h$ and the goal root position $\bm{t}_h^g$ in the input goal hand grasp label $\bm{g}$.

% \begin{equation}
%    e_{\text{mpe}} = \lVert  \bm{t}_h-\bm{t}_h^g \rVert_2
% \end{equation}

% \noindent\textbf{MRE}\quad This metric is used for measure the mean angular error between the hand root rotation of the exact grasping pose $R_h$ and the goal root rotation $R_h^g$ in the input goal hand grasp label $\bm{g}$. (using rotation matrices to represent $R_h$ and  $R_h^g$ here)

% \begin{equation}
%    e_{\text{mre}} = \text{acos}(0.5(\text{trace}(R_h {R_h^g}^T))-1)
% \end{equation}

\noindent\textbf{MPE (cm)}\quad This metric is used to measure the mean joint position error between the joint position $\bm{j}^r$ of the exact grasping pose and the joint angels $\bm{j}^g$ computed from the input goal hand grasp label $\bm{g}$ using forward kinematics. $J$ is the number of joints. Note we only calculate the MPE for the success grasp.

\begin{equation}
   e_{\text{mpe}} = \frac{1}{J} \sum_{}^{}{\lVert \bm{j}^r-\bm{j}^g \rVert_2} 
\end{equation}

\begin{figure}[t]
    \centering
    \includegraphics[width=0.92\linewidth]{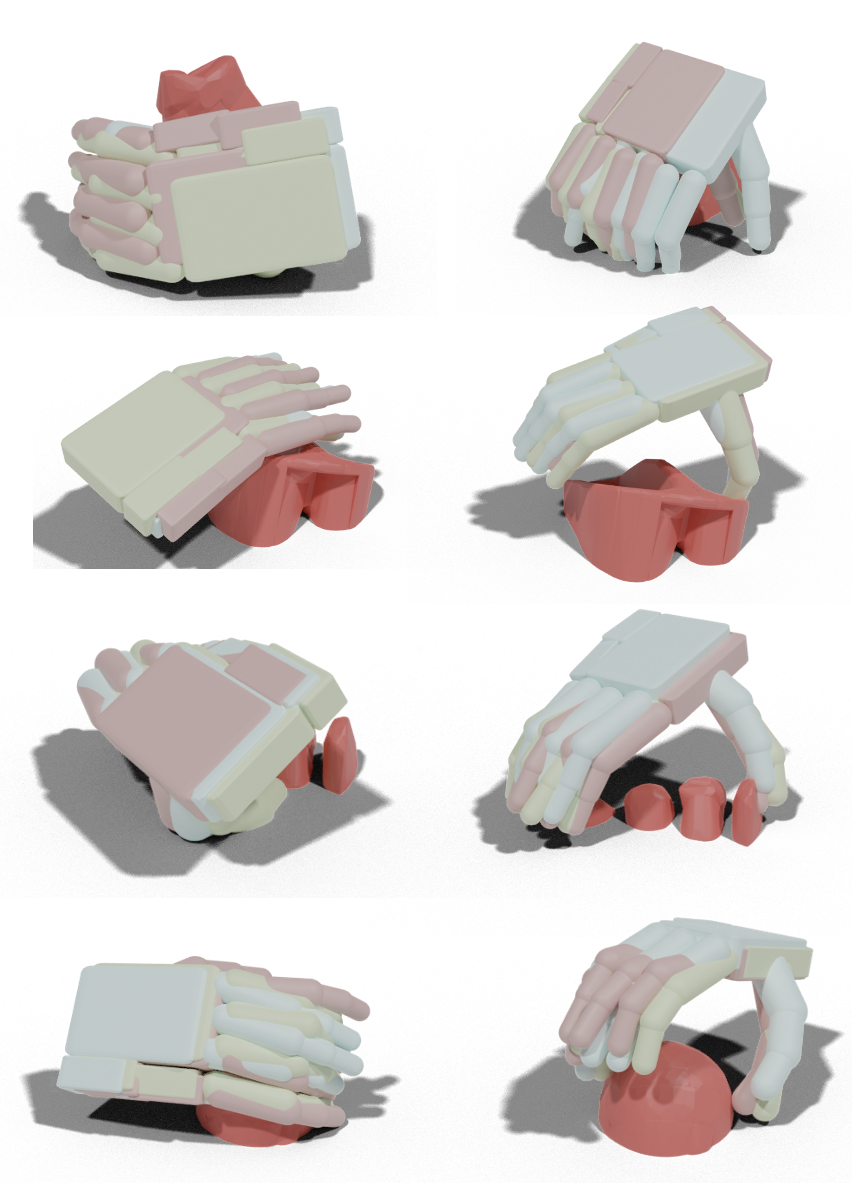}
    \caption{\textbf{Diverse grasp proposals}. Here we show the diversity of our grasping pose samples. On the same row, the objects are the same, and in the same picture, the hand root rotation is also the same. It can be shown that both the rotation samples from GraspIPDF and the translation and joint angles samples from GraspGlow have high diversity. }
    \label{fig:proposal_diversity}
\end{figure}

\section{Additional Results and Analysis}
\label{sec:add-result}

\begin{table}[!t]\small
    \centering
    \begin{tabular}{@{}l|c}
    \toprule
        Method &  Log-likelihood \\
        \hline \hline
        ReLie~\cite{falorsi2019reparameterizing}  & -1.540 \\
        ProHMR~\cite{kolotouros2021probabilistic}  & -1.710 \\
         ours (R + GL) & \textbf{10.908} \\
        \bottomrule
    \end{tabular}
    \caption{\textbf{Comparison on Log-likelihood of ground truth grasps.} R: GraspIPDF, GL: GraspGlow. Note that the outputted probability of flow in baselines on $\SO3$ is unnormalized and that we generate uniform grids using the method described in \cite{doi:10.1177/0278364909352700} and approximate the normalizing constant similar to IPDF~\cite{implicitpdf2021}.   }
    \label{tab:NLL-Table}
\end{table}

This section contains extended results of the experiment depicted in Sec.~4. 

\subsection{Goal-conditioned Dexterous Grasping Policy Results}

\noindent\textbf{Goal-conditioned vs. Non-Goal-conditioned} 
\quad We also conducted experiments in a non-goal-conditioned setting, which is the task of grasping objects alone. We compare our results with baselines described in Sec.~\ref{sec:details_baselines}. We add MP and IBS-Grasp only in supplementary because MP cannot perform non-goal-conditioned tasks and IBS-Grasp does not have a goal-conditioned setting. To perform non-goal-conditioned tasks using our method, we simply remove the goal input and goal reward in RL training. The results are shown in Tab.~\ref{tab:SuppStaticPolicy}. In non-goal-conditioned settings, our teacher policy has the highest success rate across training and all testing data sets. \\

\noindent\textbf{Analysis of Quantitative Grasping Results}
\quad The metric MPE in Tab.~\ref{tab:SuppStaticPolicy} measures the deviation between each method's interaction ending grasp and input goal grasp, as defined in Sec.~\ref{sec:metric_details}. The results show that except for the motion planning, our method has the lowest MPE among all the RL-based methods. Though the MPE of motion planning is the lowest, it has the lowest success rate, so it is unreliable. Especially, since our generated grasping proposal on unseen object categories is noisy, simply motion planning to the goal position cannot grasp the object firmly. The gaps between hand and object in generated data lead to the minimization of the MPE of MP but this kind of low MPE is helpless since it cannot seam the gap. On the contrary, our method can modify the generated noisy grasp goal and make the grasp possible. \\

\begin{figure*}[t]
    \centering
    \includegraphics[width=0.8\linewidth]{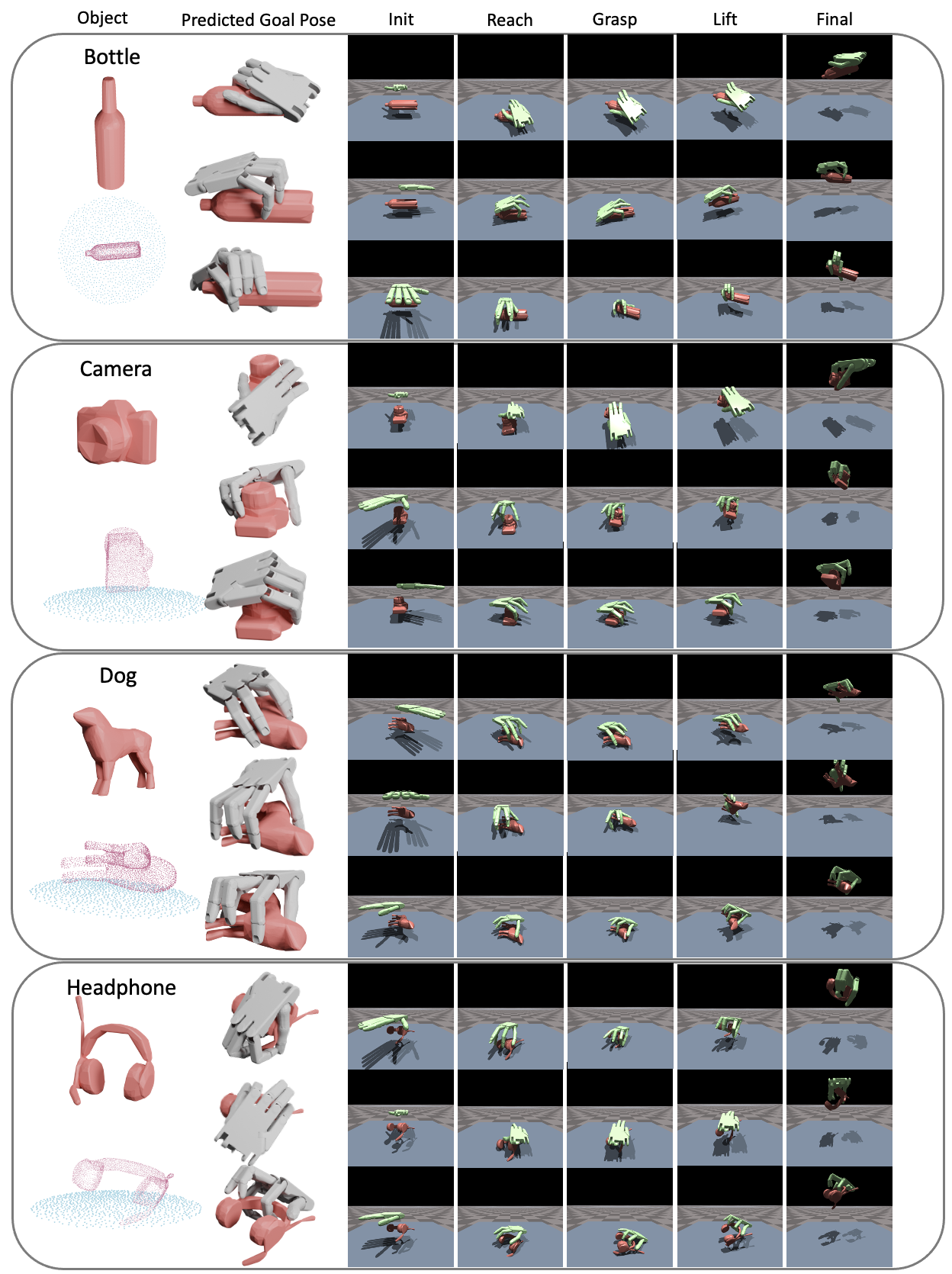}
    \caption{\textbf{Qualitative Grasping Results}. The left side includes objects and corresponding grasping poses generated using our method, and the right side is the policy-generated grasping sequence using the left corresponding part as the grasping goal. Object categories from top to bottom: bottle, camera, toy dog, and a headphone.}
    \label{fig:gallery}
\end{figure*}

\noindent\textbf{Additional Qualitative Grasping Results}
\quad We provide a qualitative demonstration of the diverse grasp proposals in Fig.~\ref{fig:proposal_diversity}. Comparing the two images of each row, one can see the diverse rotation predictions of GraspIPDF. The different hands in each image demonstrate the diverse translation and articulation predictions of GraspGlow. We also provide additional qualitative grasping results in Fig.~\ref{fig:gallery}. The left part of the figure is the visualization of the goal hand-grasping pose. For each row, the right part is the generated grasping sequence using the left part as the grasping goal, and we select four representative stages as pictures. From top to bottom, the four featured objects are a bottle, a camera, a toy dog, and a headphone. All the objects here are in the test data set, and the grasping goals are selected from our generated grasping proposals.

\end{document}